%% file: arxiv.tex
\documentclass[10pt,twocolumn,letterpaper]{article}

\usepackage{cvpr}      % To produce the REVIEW version
\usepackage{amsmath} % for mathematical environments like align
\usepackage{amssymb,bm} % for mathematical symbols
\usepackage{graphicx} 
\usepackage{makecell}
\usepackage{multirow}
\usepackage{enumitem}
\usepackage{pifont}
\usepackage{adjustbox}
\usepackage{enumitem}
\usepackage{tikz}
\usepackage{tabularx}
\usepackage{algorithm}
\usepackage[english]{babel}
\usepackage{algorithmic}
\input{preamble}

\definecolor{cvprblue}{rgb}{0.21,0.49,0.74}
\usepackage[pagebackref,breaklinks,colorlinks,allcolors=cvprblue]{hyperref}

\def\paperID{2092} % *** Enter the Paper ID here
\def\confName{CVPR}
\def\confYear{2026}

\title{EmbodMocap: In-the-Wild 4D Human-Scene Reconstruction\\ for Embodied Agents
}

\author{
Wenjia Wang$^{1*}$ \quad Liang Pan$^{1*}$ \quad Huaijin Pi$^{1}$ \quad Yuke Lou$^{1}$ \quad Xuqian Ren$^{2}$ \quad Yifan Wu$^{1}$ \\ \quad Zhouyingcheng Liao$^{1}$  \quad
Lei Yang$^{3}$ \quad Rishabh Dabral$^{4}$ \quad Christian Theobalt$^{4}$ \quad Taku Komura$^{1}$ \\[1.5mm]
{\small(*: Core contributor.)}
\\
\normalsize $^1$The University of Hong Kong \quad 
\normalsize $^2$Tampere University \\
\quad 
\normalsize $^3$The Chinese University of Hong Kong
\quad 
\normalsize $^4$Max-Planck Institute for Informatics
}

\newcolumntype{L}[1]{>{\raggedright\arraybackslash}p{#1}}
\newcolumntype{C}[1]{>{\centering\arraybackslash}p{#1}}
\newcolumntype{R}[1]{>{\raggedleft\arraybackslash}p{#1}}

\newcommand{\mypara}[1]{\noindent\textbf{#1}}

\newcommand{\newvspace}[1]{}
\begin{document}
% \maketitle

\twocolumn[{
\renewcommand\twocolumn[1][]{#1}
\maketitle
\begin{center}
    \centering
    \captionsetup{type=figure}
    \includegraphics[width=0.95\linewidth]{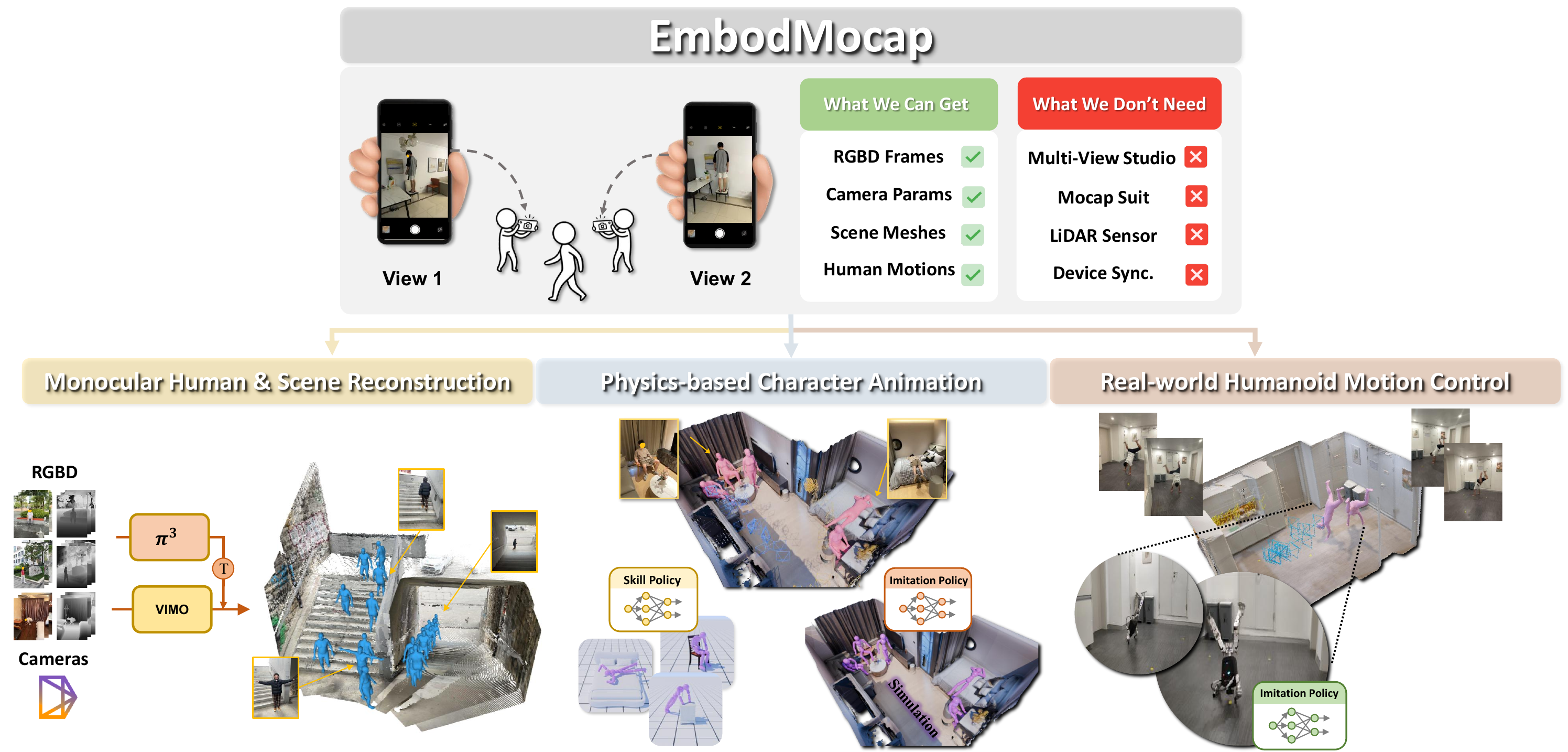}
    \captionof{figure}{Introducing \textbf{EmbodMocap}, a portable and low-cost system for simultaneous 4D human and scene reconstruction, deployable anywhere using two moving iPhones. The dataset captured by EmbodMocap benefits three crucial embodied AI tasks: monocular human \& scene reconstruction, physics-based character animation, and real-world humanoid motion control. \href{https://wenjiawang0312.github.io/projects/embodmocap}{Project page}.
    }
    \label{fig:teaser}
\end{center}
}]
% {
%   \renewcommand{\thefootnote}%
%     {\fnsymbol{footnote}}
%   \footnotetext[1]{Core contributor.}
% }

\input{sections/0_abs}    
\input{sections/1_intro}

\input{sections/2_related}
\input{sections/3_method}

\input{sections/4_exp}
\input{sections/5_conclusion}

\input{sections/6_acknowledge}
\clearpage
{
    \small
    \bibliographystyle{ieeenat_fullname}
    \bibliography{main}
}
\input{sections/X_supp}

\end{document}

%% file: preamble.tex
\definecolor{RDcolor}{rgb}{0.5, 0.1, 0.8}

%% file: sections/0_abs.tex
\begin{abstract}
Human behaviors in the real world naturally encode rich, long-term contextual information that can be leveraged to train embodied agents for perception, understanding, and acting.
However, existing capture systems typically rely on costly studio setups and wearable devices, limiting the large-scale collection of scene-conditioned human motion data in the wild.
To address this, we propose EmbodMocap, a portable and affordable data collection pipeline using two moving iPhones. 
Our key idea is to jointly calibrate dual RGB-D sequences to reconstruct both humans and scenes within a unified metric world coordinate frame.
The proposed method allows metric-scale and scene-consistent capture in everyday environments without static cameras or markers, bridging human motion and scene geometry seamlessly.
Compared with optical capture ground truth, we demonstrate that the dual-view setting exhibits a remarkable ability to mitigate depth ambiguity, achieving superior alignment and reconstruction performance over single iphone or monocular models.
Based on the collected data, we empower three embodied AI tasks: monocular human-scene-reconstruction, where we fine-tune on feedforward models that output metric-scale, world-space aligned humans and scenes; physics-based character animation, where we prove our data could be used to scale human-object interaction skills and scene-aware motion tracking; and robot motion control, where we train a humanoid robot via sim-to-real RL to replicate human motions depicted in videos. 
Experimental results validate the effectiveness of our pipeline and its contributions towards advancing embodied AI research. 
\end{abstract}

%% file: sections/1_intro.tex
% \vspace{-15pt}
\section{Introduction}
% \vspace{-5pt}
Embodied Artificial Intelligence (Embodied AI) aims to build agents that can perceive, understand, and act within real-world environments.
Progress in this field relies on datasets that capture both human motion and the surrounding 3D scene, enabling physically grounded perception and action learning.
Such scene-aware data allows modeling of realistic human–scene interactions, simulation of lifelike behaviors, and training of humanoids to operate seamlessly in complex environments.
They serve as a foundation for advancing embodied reasoning and control across robotics, virtual reality, and computer vision.

However, collecting high-quality human–scene data remains difficult.
Precise 3D motion and scene geometry cannot be automatically obtained from internet videos due to occlusions and depth ambiguity. 
Existing capture systems that provide high-quality human–scene data typically rely on multi-view camera rigs~\cite{prox, egobody}, wearable motion suits~\cite{emdb, nymeria}, or LiDAR scanners~\cite{sloper4d, rich}, which are costly, complex, and limited to controlled studio environments.  
These constraints hinder scalable and scene-aware data acquisition, limiting the ability of embodied AI models to learn from natural human behavior in diverse indoor and outdoor environments.

In this paper, we propose EmbodMocap, an efficient and affordable framework for capturing metrically accurate 4D human and scene using only two iPhones.
Our key idea is to jointly calibrate and optimize dual RGB-D inputs to reconstruct both humans and scenes within a unified world coordinate frame.
Specifically, we first reconstruct the static scene from a single RGB-D sequence to define the world scale, then capture synchronized dual-view RGB-D videos of human motion, and finally perform geometric alignment and motion optimization to recover world-anchored human poses.
In contrast to existing systems that rely on multi-camera rigs or wearable sensors, our approach achieves high-quality, scene-consistent reconstruction using only moving consumer devices.
This design enables scalable, in-the-wild data collection that preserves precise human motion and authentic scene context, supporting realistic human–scene interaction modeling for embodied AI research.

Based on the data collected with EmbodMocap, we demonstrate the reliability and versatility of our capture pipeline through three representative applications.
The first application verifies geometric consistency, where we fine-tune reconstruction models to jointly recover humans and scenes in world coordinates.
The second validates physical realism, showing that the captured motions enable scalable training of physics-based character skills and scene-aware motion tracking.
The third demonstrates embodied transferability, where our data support humanoid robot training through a sim-to-real motion tracking framework~\cite{deepmimic,liao2025beyondmimic}.
These results highlight that EmbodMocap enables scalable and physically grounded data acquisition for embodied AI.

In summary, our contributions are:
\begin{itemize}
    \item \textbf{EmbodMocap}: A portable capture framework that jointly calibrates and optimizes dual moving RGB-D cameras (iPhones) to reconstruct metrically accurate, world-anchored human motions and static scenes without multi-camera setups, mocap suits, or controlled environments.
    \item \textbf{A multi-modal dataset}: A collection of high-quality, scene-aware human motion data captured with EmbodMocap across diverse real-world environments, enabling scalable training for embodied AI.
\end{itemize}

We validate the effectiveness of our method and dataset through experiments in monocular human-scene reconstruction, physics-based character animation, and sim-to-real humanoid control, demonstrating their utility across key embodied AI tasks.

% In summary, our contributions can be summarized as follows:
% \begin{itemize}
%     \item We introduce EmbodMocap, a portable, affordable, scalable, and accessible data collection pipeline that produces high-quality multi-modal data for embodied AI applications, lowering the barrier to embodied AI research and enabling new real-world applications and future advances in the field.
%     \item 
%     \item We demonstrate the effectiveness of our capture pipeline on three key embodied AI tasks: monocular human-scene reconstruction, physics-based character animation, and real-world humanoid motion control.
%     % \item We introduce EmbodMocap, a portable and affordable data collection pipeline that produces high-quality multi-modal data for embodied AI applications.
%     % \item We validate our capture pipeline’s effectiveness across three key embodied AI tasks: monocular human-scene reconstruction, physics-based character animation, and real-world humanoid motion control.
%     % \item We provide a scalable and accessible solution that lowers the barrier for embodied AI research, opening new possibilities for real-world applications and further advancements in the field. 
% \end{itemize}

%% file: sections/2_related.tex
\section{Related Work}
\input{tables/dataset_comparision}
\mypara{Datasets for 4D Human \& Scene Capture.}
Early motion datasets, such as AMASS~\cite{amass, lafan}, focus on pure human motion, unifying multiple motion capture sources into a large-scale repository. While invaluable for studying human motion, these datasets lack the 3D scene context essential for understanding human–scene interactions. Recent 4D datasets, like PROX~\cite{prox}, RICH~\cite{rich}, and EgoBody~\cite{egobody}, combine scanned 3D scenes with motion capture using multi-view camera systems, while EMDB~\cite{emdb} and SPLOPER4D~\cite{sloper4d}, employ IMUs or electromagnetic sensors for motion recording in large-scale environments. Nymeria~\cite{nymeria} extends this further with Project Aria glasses and optical marker-based systems for wide-area motion capture.
However, these approaches face notable limitations: marker-based and multi-camera systems are expensive and restricted to small studio environments, while IMU and EM-based methods, though more flexible, require extensive manual alignment and post-processing to synchronize motion with 3D scenes. And the wearable devices will influence the human appearance in RGB images. In contrast, our approach uses minimal equipment, operates in diverse environments without static camera setups, and avoids wearable devices, preserving the naturalness of RGB images for authentic human–scene interaction capture. Table~\ref{tab:datasets} compares these datasets.

\mypara{Monocular Human \& Scene Reconstruction.}
Early works~\cite{smplify,smplx,hmr,hmr2,vibe} on RGB-based human mesh recovery focus on reconstructing 3D pose and shape but often ignore scene context~\cite{deco} or camera information~\cite{spec, zolly}, leading to inconsistencies under camera motion. Recent methods address this by combining motion cues~\cite{glamr}, SLAM or visual odometry~\cite{slahmr,tram, wham}, and human motion priors~\cite{glamr, gvhmr} to recover global trajectories in world coordinates.

Emerging models move toward jointly reconstructing humans and 3D scenes with spatial intelligence models~\cite{vggt, dust3r}. For example, HSFM~\cite{hsfm} combines Dust3R~\cite{dust3r} with multi-view correspondence to jointly recover human meshes, scene point clouds, and camera parameters from multi-cameras. HAMSt3R~\cite{hamst3r} integrates DensePose~\cite{densepose} and multi-view scene reconstruction in one model, with an optimization to get human poses, while JOSH~\cite{josh} uses MASt3R-SLAM~\cite{mast3r-slam} and joint optimization to achieve globally consistent 4D human-scene reconstructions. Human3R~\cite{human3r} introduces a unified, feed-forward framework for online 4D human-scene reconstruction, jointly recovering multi-person SMPL-X bodies and dense scene point clouds in a global world frame from monocular videos. Crisp~\cite{crisp} presents a contact-guided Real2Sim pipeline that recovers simulatable human motion and scene geometry by fitting compact planar primitives and leveraging human-scene contact cues to hallucinate occluded interaction surfaces. This trend emphasizes the simultaneous prediction of human motion and scene geometry, which further requires multi-model data pairs with high-quality annotations. In our paper, we propose a monocular human \& scene reconstruction pipeline combined with 2 feedforward models, and finetuned it on our proposed dataset to prove the efficiency of our paired data.

\mypara{Training Humanoid from Video Data.}
Recent advances in physics-based animation and reinforcement learning enable humanoid agents to perform realistic and physically consistent motions using control policies learned from marker-based motion capture data. These methods have shown strong realism in tasks like motion tracking~\cite{deepmimic,phc}, locomotion~\cite{amp,ase,pulse}, and human–scene interaction~\cite{tokenhsi,sims}, and have been extended to real-world applications in motion tracking~\cite{h2o,exbody2,asap}, locomotion~\cite{hover}, and scene interaction~\cite{omnih2o,homie}.
However, marker-based methods require dedicated studios, expensive hardware, and extensive manual effort, making them costly and hard to scale. Adapting captured motions to new scenes or robot morphologies also demands complex retargeting and re-simulation. To address this, recent works like VideoMimic~\cite{videomimic}, ASAP~\cite{asap}, and HDMI~\cite{hdmi} train humanoid control directly from in-the-wild video data. By using monocular motion capture methods such as TRAM~\cite{tram} and GVHMR~\cite{gvhmr}, they estimate human motion from videos and retarget it to virtual humanoids for training in physical simulators. This video-driven paradigm leverages diverse real-world data but struggles with capturing complex skills or scene geometries due to occlusion and depth ambiguities. In this paper, we propose a method for high-precision human motion and scene reconstruction that overcomes these limitations. 

%% file: tables/dataset_comparision.tex
\begin{table*}[ht]
\centering
\caption{Comparison of 4D Human~\&~Scene datasets based on different features.}
% \vspace{-7pt}
\resizebox{0.99\linewidth}{!}{
\begin{tabular}{l|C{50pt}|C{110pt}C{80pt}C{80pt}C{80pt}C{80pt}|C{30pt}C{50pt}C{30pt}}
\toprule[1.5pt]
\multirow{2}{*}{\textbf{Datasets}} & \multirow{2}{*}{\textbf{Publication}} & \multicolumn{5}{c|}{\textbf{Device}} & \multicolumn{3}{c}{\textbf{Outcome}} \\ 
&  & Mocap Suit & Scanner & Static Cam. & Dyna. Cam. & Total Cost(\$) & Mesh & Dyna.Anno. & Outdoor \\ \midrule[1pt]
PROX~\cite{prox} & ICCV2019 & - & Structure Sensor & Kinetic-One &  -& 2K & \ding{51} & \ding{55} &  \ding{55}\\ \midrule
RICH~\cite{rich} &  CVPR 2022 & - & Leica RTC360 & 6-8$\times$Cameras & 1$\times$Camera & 20K+ & \ding{51} & \ding{51} & \ding{51} \\ \midrule

EgoBody~\cite{egobody} & ECCV2022 & - & 1$\times$IPhone & 5$\times$Azure Kinect & Hololens2 & 9K & \ding{51} & \ding{51} & \ding{55}\\ \midrule

SLOPER4D~\cite{sloper4d} & CVPR2023 & Noitom PN+NUC11 & Ouster-os1 LiDAR & - & DJI-Action2+TLS & 20K & \ding{51} & 
\ding{51} & \ding{51} \\ \midrule

EMDB~\cite{emdb} &  ICCV 2023 & EM Sensors & - & - & 1$\times$IPhone & 15K & \ding{55} & \ding{51} & \ding{51} \\ \midrule
% EgoHDM~\cite{egohdm} & ToG2024 & IMU suits &  & \ding{51} & \ding{51} & 30K & \ding{51} & & \ding{51} \\ \midrule
Nymeria~\cite{nymeria}  &ECCV2024 & 2$\times$XSens+Aria Wistband & - & - & 2$\times$Project Aria & 60K+ & \ding{55} & \ding{51} & \ding{51} \\ \midrule[1pt]
EmbodMocap & -  & - & 1$\times$IPhone  & -  & 2$\times$IPhone & 1K & \ding{51} & \ding{51} & \ding{51} \\ 
\bottomrule[1.5pt]
\end{tabular}}

\label{tab:datasets}
% \vspace{-12pt}
\end{table*}

%% file: sections/3_method.tex
\section{Proposed Capture System}
\label{sec:embodmocap}
\begin{figure*}[!ht]
    \centering
    % 设置占位框的宽度和高度
    \includegraphics[width=1\linewidth]{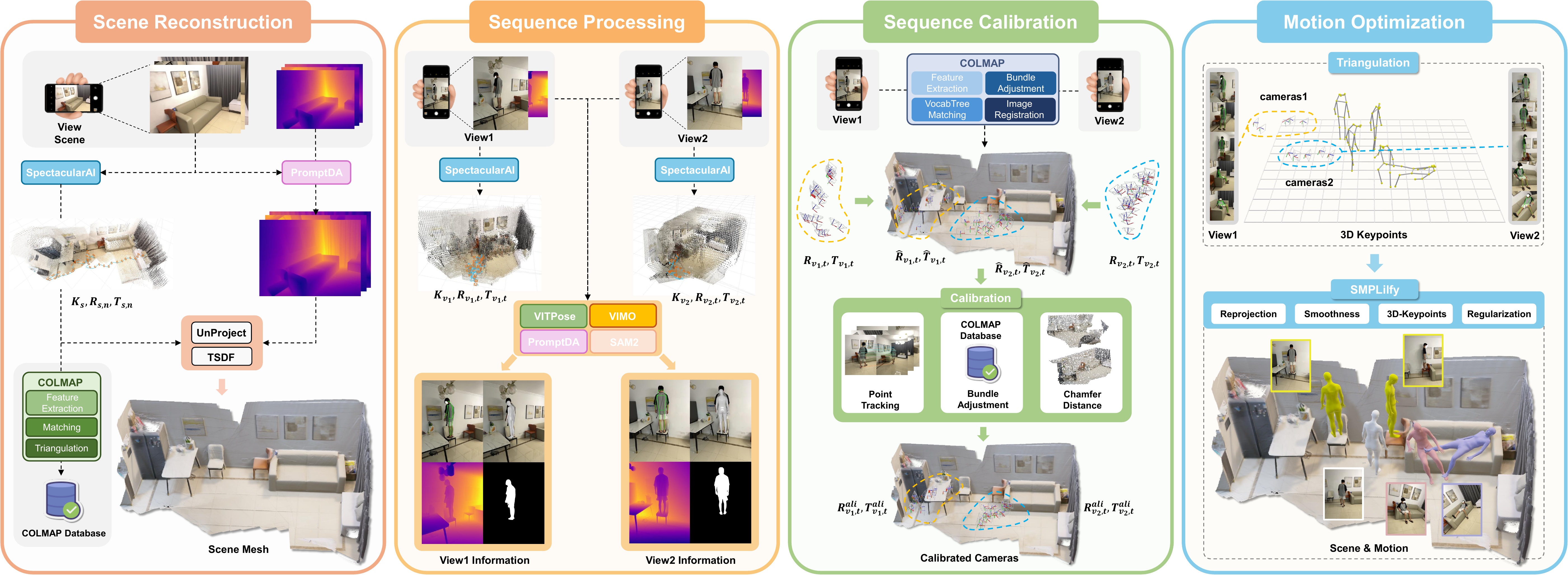}
     % \vspace{-5pt}
    \caption{EmbodMocap: We propose an affordable dataset capture and processing system. From left to right, the four stages (Stage-I to Stage-IV) illustrate our core logic: leveraging high-quality camera matrices provided by SpectacularAI~\cite{spectacularai} and aligning sequence coordinates to the scene's world frame.  For detailed explanations, please refer to ~\cref{sec:embodmocap}.
 }
 \label{fig:embodmocap}
 % \vspace{-10pt}
\end{figure*}

We aim to capture metrically accurate human motion and scene geometry using only two iPhones.  
As shown in Fig.~\ref{fig:embodmocap}, our capture process consists of four sequential stages that progressively reconstruct and align the scene, cameras, and human motion within a unified world coordinate frame.  
We first reconstruct a metrically accurate static scene and establish the world reference using a single iPhone RGB-D sequence (Sec.~\ref{sec:stage1}).  
Then, we use two synchronized iPhones to record dual-view RGB-D videos of human motion and extract per-frame camera poses and human priors with off-the-shelf perception models (Sec.~\ref{sec:stage2}).  
Next, we align the dual-view camera trajectories to the reconstructed scene through a combination of COLMAP registration and multi-view geometric optimization (Sec.~\ref{sec:stage3}).  
Finally, we refine the SMPL parameters by triangulating dual-view 2D keypoints into 3D space and optimizing human poses and translations in the world coordinate system (Sec.~\ref{sec:stage4}).

\input{sections/subsections/data_stage1}

\input{sections/subsections/data_stage2}
\input{sections/subsections/data_stage3}
\input{sections/subsections/data_stage4}

%% file: sections/subsections/data_stage1.tex
\subsection{Stage I: Scene Reconstruction}
\label{sec:stage1}
In this stage, we aim to reconstruct a metrically accurate, Z-up scene mesh that serves as the reference world coordinate system.  
We first use a single iPhone to capture an RGB-D video of the scene, along with synchronized IMU data.  
The recorded data are processed by the SpectacularAI SDK (SAI) \cite{spectacularai}, which automatically selects keyframes according to the accumulated camera translation and estimates corresponding camera parameters $(\bm{K}_s, \bm{R}_{s,n}, \bm{T}_{s,n})$ in Z-up world coordinates with metric scale.  
These trajectories establish a consistent world frame for all subsequent stages.  
Based on the recovered poses, we refine the iPhone LiDAR depth maps using PromptDA~\cite{promptda}, unproject them into 3D space, and integrate the point clouds through TSDF fusion~\cite{tsdf} to obtain a dense and metrically accurate global mesh $\mathcal{M}_g$. Note that the depth maps are truncated based on a threshold determined by the effective range of the iPhone's depth sensor. Specifically, we use a threshold of 3.5m for indoor scenes and 5m for outdoor scenes.
We further apply lightweight post-processing such as outlier removal and small-component filtering to clean the mesh.  
Finally, we extract SIFT features from the same SAI keyframes and run COLMAP \cite{colmap} with fixed camera parameters to build a sparse structure database.  
This database preserves the metric scale and serves as a reference for registering dual-view sequences in later stages.

%% file: sections/subsections/data_stage2.tex
% \vspace{-6pt}
\subsection{Stage II: Sequence Processing}
\label{sec:stage2}
After reconstructing the static scene in Stage I, we proceed to capture and process dual-view human motion sequences within the same environment.
In this stage, we use two iPhones to record synchronized RGB-D videos of a performer moving inside the reconstructed scene, with each device providing an independent camera coordinate system.
The goal is to convert these raw dual-view videos into temporally aligned and metrically consistent per-frame human and camera information, which will serve as the foundation for subsequent calibration and motion optimization.

Firstly, we use SAI to obtain per-frame calibrated cameras for each view. Let $v$ denote the view index ($v \in \{v_1, v_2\}$), and let $t$ denote index time. For each view independently, SAI provides intrinsics and extrinsics $(\bm{K}_{v}, \bm{R}_{v,t}, \bm{T}_{v,t})$ for every decoded frame $\bm{I}_{v,t}$ in the native coordinate system of that view. 

Next, we extract human-related information using several off-the-shelf models: (i) YOLO~\cite{yolo} for person detection and proposal pruning; (ii) ViTPose~\cite{vitpose} for 2D human keypoints with confidence scores; (iii) SAM2~\cite{sam2} for person segmentation masks; (iv) PromptDA~\cite{promptda}  to refine dual-view depths; and (v) VIMO~\cite{tram} for camera space SMPL parameters.
Finally, we employ a laser pointer cue for frame-level synchronization between the two camera streams.
By identifying the frame index where the laser dot disappears, we temporally align both videos and slice all associated image, depth, and parameter data accordingly.
This process yields synchronized dual-view RGB-D sequences with calibrated camera trajectories and per-frame human priors, providing clean inputs for subsequent sequence calibration.

%% file: sections/subsections/data_stage3.tex
\subsection{Stage III: Sequence Calibration}
\label{sec:stage3}
\vspace{0pt}

After obtaining the static scene reconstruction in Stage~\ref{sec:stage1} and the dual-view camera trajectories in Stage~\ref{sec:stage2}, the next step is to align all coordinate systems into a unified world frame.  
At this point, we have three separate coordinate systems: one for the reconstructed scene and two for each iPhone camera trajectory estimated by SAI.  
Since the dual-view coordinate systems differ from the scene coordinate system only by rigid transformations, our goal is to optimize these 2 rigid transformations to unify the dual-view coordinates into the same metric, gravity-aligned world frame. 
The optimization process is sensitive to the initial values; therefore, it is necessary to first obtain a good initial estimate for the rigid transformations.

\mypara{Get Initial Transformation from COLMAP.}  
We register each dual-view sequence to the sparse COLMAP model constructed in Stage~\ref{sec:stage1} using the known intrinsics $K_v$ and background-only SIFT features $\mathcal{F}_v$, extracted from images with human regions removed.  
Matches are established through a trained vocabulary tree~\cite{vocabtree}, and images are registered against the sparse COLMAP model to obtain COLMAP camera poses $(\hat{\bm{R}}_{v,t}, \hat{\bm{T}}_{v,t})$ in the same metric, gravity-aligned world coordinates as the scene.  

To obtain the initial rigid transformation aligning the SAI camera trajectories ${\bm{T}_{v,t}}$ with their COLMAP counterparts ${\hat{\bm{T}}_{v,t}}$, we solve for an offset transformation $(s^{\mathrm{off}}, \bm{R}^{\mathrm{off}}, \bm{T}^{\mathrm{off}})$ by minimizing:
\vspace{0pt}
{
\begin{equation}\label{eq:procrustes_stage3}
\min_{s^{\mathrm{off}}, \bm{R}^{\mathrm{off}}, \bm{T}^{\mathrm{off}}}
\sum_{t=1}^{N} \big\lVert \hat{\bm{T}}_{v,t} - (s^{\mathrm{off}} \bm{R}^{\mathrm{off}} \bm{T}_{v,t} + \bm{T}^{\mathrm{off}}) \big\rVert_2^2,
\end{equation}
}
\noindent where $N$ is the number of frames. After centering the trajectories, we solve this minimization problem using singular value decomposition (SVD).

For gravity alignment, $\bm{R}^{\mathrm{off}}$ is constrained to rotations about the $z$-axis, ensuring proper alignment of SAI trajectories with the COLMAP coordinate system.

\mypara{Calibration via Multiple Constraints.}
While the rigid transformations obtained in the previous step provide coarse alignment between the two camera trajectories and the reconstructed scene, this initialization alone is not sufficient to achieve accurate synchronization and metric consistency.  
To further refine the calibration, we jointly optimize all alignment parameters by introducing multiple geometric and photometric constraints across views. Specifically, we optimize the per-view global offsets $R_v^{\mathrm{off}}$ (constrained to $z$-axis rotations) and $T_v^{\mathrm{off}}$, using the initial alignment as the starting value. The aligned camera extrinsics are:

\vspace{0pt}
{
\begin{equation}
\bm{R}_{v,t}^{\mathrm{ali}} = \bm{R}_v^{\mathrm{off}} \bm{R}_{v,t}, \quad
\bm{T}_{v,t}^{\mathrm{ali}} = \bm{R}_v^{\mathrm{off}} \bm{T}_{v,t} + \bm{T}_v^{\mathrm{off}}.
\label{eq:pose_transform_stage3}
% \vspace{0pt}
\end{equation}
}

The optimization minimizes a composite loss of point tracking loss, Chamfer distance, and bundle adjustment loss to ensure spatial consistency between views and the global reconstruction.
% \vspace{0pt}
{
\begin{equation}
\mathcal{L}_{\mathrm{calib}} = \lambda_{\mathrm{track}} \mathcal{L}_{\mathrm{track}} 
+ \sum_{v} \lambda_{\mathrm{ch}} d_{\mathrm{Chamfer}} 
+ \sum_{v} \lambda_{\mathrm{ba}} \mathcal{L}_{\mathrm{ba},v}.
\label{eq:calib_objective_stage3}
 \vspace{0pt}
\end{equation}}
\noindent
where each loss is defined in the rest of this seciton. 

Through VGGT tracking, a subset of keyframes is selected, yielding accurate dual-view pixel tracking results in the human masks region. The tracked human surface 2D pixel coordinates $\bm{q}_{v,t}^{(i)}$, along with their corresponding depth values $d_{v,t}^{(i)}$, are back-projected into the world frame:
% \vspace{0pt}
{
\begin{equation}
\bm{Q}_{v,t}^{(i)} = d_{v,t}^{(i)} \bm{R}_{v,t}^{\top \mathrm{ali}} \bm{K}_v^{-1} 
\begin{bmatrix}
\bm{q}_{v,t}^{(i)} \\ 1
\end{bmatrix} + \bm{R}_{v,t}^{\top \mathrm{ali}} \bm{T}_{v,t}^{\mathrm{ali}},
\label{eq:back_projection}
\end{equation}
}

To enforce track consistency between views, the following loss is minimized:
% \vspace{0pt}
{
\begin{equation}
\mathcal{L}_{\mathrm{track}} = \frac{1}{\sum_{v,t} |\mathcal{Q}_{v,t}|}
\sum_{t} \sum_{i} \tilde{w}_{t}^{(i)}
\big\| \bm{Q}_{1,t}^{(i)} - \bm{Q}_{2,t}^{(i)} \big\|_2^2,
\label{eq:track_loss}
 \vspace{0pt}
\end{equation}
}

\noindent
where $\bm{Q}_{1,t}^{(i)}$ and $\bm{Q}_{2,t}^{(i)}$ are the 3D back-projected coordinates of the $i$-th point from view $1$ and view $2$, respectively. The weights $\tilde{w}_t^{(i)}$ are used to control the contribution of each point based on its tracking confidence.
Here $\tilde{w}_{t}^{(i)} = \min(w_{1,t}^{(i)}, w_{2,t}^{(i)})$ combines the VGGT confidence scores for the same point across views.
The Chamfer distance term $d_{\mathrm{Chamfer}}$ aligns local pointclouds $\bm{\mathcal{P}}_v~(v\in\{v_1, v_2\})$ with the global reconstruction $\bm{\mathcal{P}}_{\mathrm{g}}$ sampled from $\mathcal{M}_\mathrm{g}$ in \cref{sec:stage1}, where $\bm{\mathcal{P}}_v$ is obtained by reconstructing the scene using the method from \cref{sec:stage1} with humans cropped by masks. The Chamfer distance is formally defined as:  
\vspace{0pt}
{
\begin{align}
d_{\mathrm{Chamfer}}(\bm{\mathcal{P}}_v, \bm{\mathcal{P}}_{\mathrm{g}})
&= \frac{1}{|\bm{\mathcal{P}}_v|} 
\sum_{\bm{p}_v \in \bm{\mathcal{P}}_v} 
\min_{\bm{p}_{\mathrm{g}} \in \bm{\mathcal{P}}_{\mathrm{g}}} 
\|\bm{p}_v - \bm{p}_{\mathrm{g}}\|_2^2 \nonumber \\
&\quad + \frac{1}{|\bm{\mathcal{P}}_{\mathrm{g}}|} 
\sum_{\bm{p}_{\mathrm{g}} \in \bm{\mathcal{P}}_{\mathrm{g}}} 
\min_{\bm{p}_v \in \bm{\mathcal{P}}_v} 
\|\bm{p}_{\mathrm{g}} - \bm{p}_v\|_2^2.
\label{eq:chamfer_stage3}
\end{align}
}
\vspace{0pt}

Finally, $\mathcal{L}_{\mathrm{ba},v}~(v\in\{v_1, v_2\})$ ensures reprojection consistency for persistent matches, where the points are obtained from COLMAP image registration:  
\vspace{0pt}
{
\begin{equation}
\mathcal{L}_{\mathrm{ba},v}
= \frac{1}{|M_v|}\sum_{(t,j) \in M_v}
\big\| \bm{x}_{v,t,j} - \pi(\bm{K}_v, \bm{R}_{v,t}^{\mathrm{ali}}, \bm{T}_{v,t}^{\mathrm{ali}}, \bm{X}_j) \big\|_2^2.
\label{eq:ba_term_stage3}
\end{equation}
}
\vspace{0pt}

We solve Eq.~\eqref{eq:calib_objective_stage3} using the Adam~\cite{adam} optimizer with gradient clipping. For yaw-only updates, $R_v^{\mathrm{off}}$ is parameterized by a single z-axis angle to preserve gravity alignment.

%% file: sections/subsections/data_stage4.tex
\subsection{Stage IV: Motion Optimization}
\label{sec:stage4}
After obtaining calibrated dual-view trajectories and a unified scene coordinate system in Stage~\ref{sec:stage3}, we further refine the human reconstruction results to achieve accurate and temporally consistent body motions in the world frame.  
At this stage, both camera poses and scene geometry are fixed, allowing us to focus on optimizing the human parameters.  
We first triangulate dual-view 2D keypoints into world-space 3D keypoints, which serve as reliable geometric constraints across views. 
Then, we optimize the SMPL parameters using these triangulated 3D keypoints to recover precise body poses and translations under the unified world coordinate system.  

\mypara{3D Keypoint Triangulation.}  
To triangulate the 3D keypoints $\bm{Y}_{t,j}$ from their 2D projections $\{y_{v,t,j}\}$, we estimate the 3D position by minimizing the weighted reprojection error across all views:
%\vspace{-5pt}
{
\begin{equation}
\min_{\bm{Y}_{t,j}} \sum_{v=1}^V c_{v,t,j} \big\|\bm{y}_{v,t,j} - \bm{P}_v \bm{Y}_{t,j} \big\|_2^2,
\end{equation}
\label{eq:triangulation_objective}
}
%\vspace{-5pt}
where $\bm{P}_v = \bm{K}_v [\bm{R}_{v,t} \ | \ \bm{T}_{v,t}]$ is the camera projection matrix for the $v$-th view.
The problem can be formulated as a weighted least squares optimization.  Using SVD, $\bm{Y}_{t,j}$ is obtained as the right singular vector corresponding to the smallest singular value of $\bm{A}$.

\mypara{World-Space SMPLify.}  
Start from initial shape  $\bm{\beta}_0$ and body pose $\bm{\theta}_t^\mathrm{b,0}$ in \cref{sec:stage2}, our World Frame SMPLify~\cite{smpl} jointly optimizes shape $\bm{\beta} \in \mathbb{R}^{10}$, per-frame pose $\bm{\theta}_t = \{\bm{\theta}_{t}^\mathrm{g}, \bm{\theta}_t^\mathrm{b}\} \in \mathbb{R}^{72}$ and root translation $\bm{\gamma}_t \in \mathbb{R}^3$ by minimizing: 
% \vspace{-5pt}

{
\begin{align}
&\mathcal{L}_{\mathrm{SMPLify}} 
= \mathcal{L}_{3\mathrm{D}} 
+ \mathcal{L}_{\mathrm{smooth}} + \mathcal{L}_{\mathrm{prior}} + \mathcal{L}_{\mathrm{reproj}}
\label{eq:smpl_objective} 
\end{align}
}
We use a two-stage optimization phase to ensure the smoothness and alignment with the original dual views. For the first stage, we only fit the body shape and transition, and for the second stage we fit all the parameters.

%% file: sections/4_exp.tex
\section{Evaluation}

In this section, we aim to prove the effectness of our optimization pipeline. We will first ablate different loss functions of the pipeline in \cref{sec:abl}, then compare ours with the monocular model, single-view only and optical captured ground truth.

\input{sections/subsections/exp4_abl}
\input{sections/subsections/exp5_optical}

\section{Downstream Tasks}
In this section, we validate our capture pipeline’s effectiveness across three key applications. In \cref{sec:exp1}, we propose a monocular human~\&~scene reconstruction pipeline and finetune it with our captured RGBD, cameras, and SMPL annotations. In \cref{sec:exp2}, we train several human-object interaction skills and scene-aware motion tracking with our captured motion \& scene. In \cref{sec:exp3}, we train a humanoid in simulator and deploy it to real-world robot.

\subsection{Monocular Human~\&~Scene Reconstruction}
\label{sec:exp1}
\input{sections/subsections/exp1_mocap}

\subsection{Physics-based Character Animation}
\label{sec:exp2}

\subsubsection{Human Object Interaction Skill Training}
\input{sections/subsections/exp2_skill}

\subsubsection{Scene-aware Motion Tracking}
\input{sections/subsections/exp2_track}

\subsection{Real-world Humanoid Robot Control}
\label{sec:exp3}

\input{sections/subsections/exp3_robo}

%% file: sections/subsections/exp4_abl.tex
\subsection{Ablation Study on Loss Functions}
\label{sec:abl}
\mypara{Ablation on dataset optimization.}
We conduct an ablation study on four core loss functions that significantly influence performance during data optimization, as described in main paper. These loss functions include tracking loss, Chamfer distance, reprojection loss, smoothness loss and kp3d loss. 
To evaluate the performance under different optimization settings, we employ four metrics. First, \textbf{IoU(Intersection over Union)} measures the overlap between the rendered SMPL mask and the SAM2~\cite{sam2} mask. Second, \textbf{Reproj} evaluates the pixel error between the reprojected SMPL joints and the 2D keypoints detected by VITPose~\cite{vitpose}. Third, \textbf{Depth} error is computed as the mean squared error (MSE) between the rendered depth from SMPL parameters and the sensor depths refined by PromptDA~\cite{promptda}. Finally, \textbf{Jitter} is quantified using the same temporal foot skating metric as MotionVAE~\cite{motionvae}. 
All metrics are averaged across all sequences and views to ensure a robust evaluation.

The $\mathcal{L}_{track}$ effectively stitches the two views together, significantly improving the overall reconstruction performance, making it highly impactful on the final results.
The $\mathcal{L}_{kp3d}$ provides 3D joint positions of the human body, and compared to the reprojection loss, it eliminates the issue of depth ambiguity, thus playing a critical role in the overall performance.
\input{tables/exp_mocap_optim}

%% file: tables/exp_mocap_optim.tex
\begin{table}[h]
\centering
 % \vspace{-2pt}
 \caption{The performance of different optimization settings.}   
 % \vspace{-6pt}
\scalebox{0.58}{
\begin{tabular}
{C{30pt}C{35pt}C{30pt}C{35pt}C{25pt}C{35pt}C{40pt}C{35pt}C{35pt}}
\toprule[1.5pt]
$\mathcal{L}_{track}$ & $\mathcal{L}_{chamfer}$ & $\mathcal{L}_{reproj}$ & $\mathcal{L}_{smooth}$ & $\mathcal{L}_{kp3d}$  & IoU(\%)$\uparrow$ & Reproj$\downarrow$ & Depth$\downarrow$ & Jitter$\downarrow$ \\ \midrule
   \ding{55}      &      \ding{51}  &      \ding{51}      & \ding{51}      & \ding{51}          & 54.3        &  44.2 &  2.372 & 0.0371     \\ \midrule
      \ding{51}      &      \ding{55}  &      \ding{51}      & \ding{51}     & \ding{51}           &  \underline{72.5}         &   10.9 &  0.081 & 0.0131    \\ \midrule
   \ding{51}    &      \ding{51}    &      \ding{55}        &     \ding{51}    & \ding{51}          &      72.3     &     11.1 & \underline{0.079} & 0.0130   \\ \midrule
      \ding{51}  &      \ding{51}   &    \ding{51}       &  \ding{55}      & \ding{51}             &    72.1    &  \underline{10.4} & 0.087 & 0.0160       \\ \midrule  
       \ding{51}  &      \ding{51}  &  \ding{51}         &   \ding{51}     & \ding{55}        &  59.3       & 20.4 & 0.609 & \textbf{0.0126}  \\ \midrule[1pt]
       \ding{51}  &      \ding{51}  &  \ding{51}         &   \ding{51}     & \ding{51}        & \textbf{73.0} & \textbf{9.3} & \textbf{0.078} & \underline{0.0128}  \\
       \bottomrule[1.5pt]
 \end{tabular}}
 % \vspace{-5pt}
      
\end{table}

%% file: sections/subsections/exp5_optical.tex
\subsection{Comparison on Capture Methods}
\label{sec:optical}
\begin{figure}[ht]
    \centering
    % \vspace{-10pt}
    \includegraphics[width=1\linewidth]{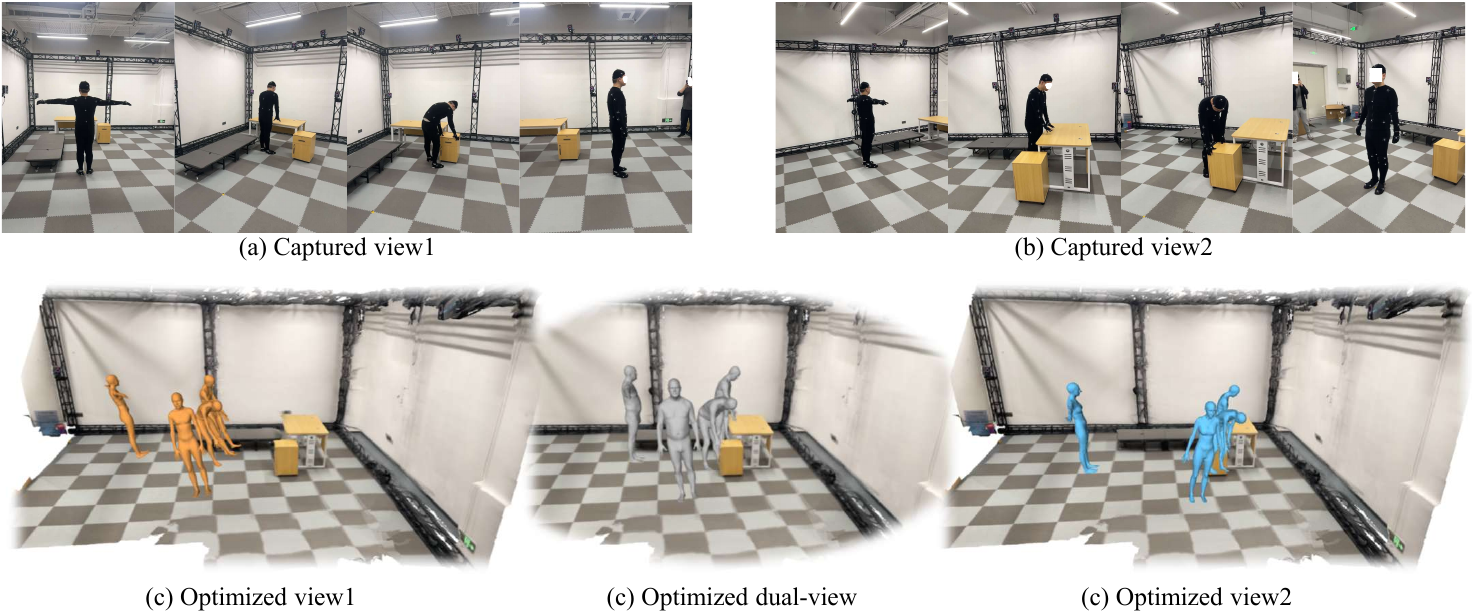}
    \caption{Our dual view \vs single view results in optical studio.}
    % \vspace{-5pt}
    \label{fig:optical}
\end{figure}

\noindent\textbf{Direct comparison in optical mocap studio.}  
To evaluate the accuracy of dual view capture system, we set up furniture in a mocap studio and use a Vicon system to capture ground truth human motion. Two photographers record dual-view videos of the actor with iPhones, while the actor performs basic motions (see \cref{fig:optical}, zoom in). We record 5 sequences of one participant with 9420 frames in total. We compare the errors against optical mocap GT of: monocular model GVHMR, our dual-view optimization, and our single-view version (v1 and v2). For the single-view version, we calibrate the actor coordinates to the scene coordinates system using COLMAP and optimize the motion with reprojection, smooth, and prior losses. The optical mocap results are fitted to  SMPLX parameters by Mosh~\cite{mosh} and synchronized to dual-view parameters with foot contact keyframs. Results are compared in chunk sizes of 100, 500, and 1000. Our dual-view method outperforms the monocular model and single-view optimization by a large margin. As the chunk length increases, our advantage becomes increasingly evident
(see \cref{tab:optical}).

\begin{table}[h]
\centering
\caption{Comparision among monocular model, single view optimization, with dual view optimization(ours)}
% \vspace{-5pt}
\resizebox{0.48\textwidth}{!}{
\begin{tabular}
{c|C{55pt}C{50pt}|C{55pt}C{50pt}|C{55pt}C{50pt}|C{40pt}}
\hline
\multirow{2}{*}{\textbf{Method}} & \multicolumn{2}{c|}{chunk=100} & \multicolumn{2}{c|}{chunk=500} & \multicolumn{2}{c|}{chunk=1000} & \multirow{2}{*}{RTE$\downarrow$} \\ \cline{2-7}
                                 & WA-MPJPE$\downarrow$ & W-MPJPE$\downarrow$ & WA-MPJPE$\downarrow$ & W-MPJPE$\downarrow$ & WA-MPJPE$\downarrow$ & W-MPJPE$\downarrow$ & \\ \hline
GVHMR                            & 66.56    & 123.44  & 124.61   & 333.34  & 179.47   & 593.79  & 1.85 \\
Single-View V1                   & 124.68   & 218.22  & 233.06   & 489.11  & 297.83   & 768.31  & 2.71 \\
Single-View V2                   & 108.31   & 211.83  & 231.41   & 357.22  & 338.42   & 762.80  & 3.65 \\
Dual View                        & \textbf{56.61}    & \textbf{72.86}   & \textbf{76.90}    & \textbf{99.75}   & \textbf{119.45}   & \textbf{169.11}  & \textbf{1.13} \\ \hline
\end{tabular}
}
\label{tab:optical}
% \vspace{-12pt}
\end{table}

\noindent\textbf{The advantage of dual-view over single-view} lies in two key aspects: 1) dual-view effectively addresses occlusion and self-occlusion of body joints, 2) it handles the challenging alignment of actor motion coordinates to the scene coordinates. 
The COLMAP estimates the camera locations for the images but suffers from depth ambiguity in the camera's facing direction. Using a single iPhone results in large errors in the depth direction. In contrast, using two iPhones enables pixel-wise dense correspondence (see \cref{eq:track_loss}), which ensures the rigid transformation between the two cameras during the optimization, and resolves the depth ambiguity in each view.
{\textit{\textbf{This enables a good localization of human trajectories in the scene coordinate system automatically.}}
Our dual view could achieve a calibration accuracy to the scene of about 5cm~(human touching table in the figure), while the single view is over 30cm, measured in MeshLab by putting markers on the ground for the actor's start and end positions.

%% file: sections/subsections/exp1_mocap.tex
\mypara{Motivation.}
We propose a data scheme combining RGBD data from dynamic cameras with camera and human motion parameters to train monocular human and scene reconstruction models. As no feedforward model exists, we establish a baseline using $\pi^3$\cite{pi3} for SLAM and VIMO\cite{tram} for metric-scale human motion reconstruction from monocular videos.

\mypara{Implementation.}
To process long sequences, videos are divided into overlapping chunks, with $\pi^3$ estimating camera parameters and local point maps per chunk. Adjacent chunks are aligned using Procrustes alignment, and scale/transformations are recursively applied for global consistency. Metric scale is determined as the median ratio of SMPL to $\pi^3$ depth values. SMPL predictions are then transformed to metric world space. For details, refer to Supp. Mat.
We fine-tuned two $\pi^3$ variants~\cref{tab:emdb} by adding LoRA~\cite{lora} layers to the camera and point decoders, supervised with the original $\pi^3$ loss. For VIMO, we froze the encoder and finetuned the decoder with MSE loss on SMPL parameters. A human mask was used to limit supervision to the human region due to our dataset's smaller range.

\mypara{Metrics.}
We evaluate motion and trajectory accuracy on global coordinates using EMDB (subset 2)~\cite{emdb}, featuring extended sequences with ground-truth trajectories and meshes. Consistent with prior work~\cite{tram, wham}, each sequence is split into 100-frame chunks, and 3D joint errors are measured using W-MPJPE (aligning the first two frames) and WA-MPJPE (aligning the entire segment), both in millimeters. Additionally, Root Translation Error (RTE) is reported as a percentage (\%), normalized by total displacement after rigid alignment (excluding scaling).

\begin{figure}[!ht]
    \centering
\includegraphics[width=0.98\linewidth]{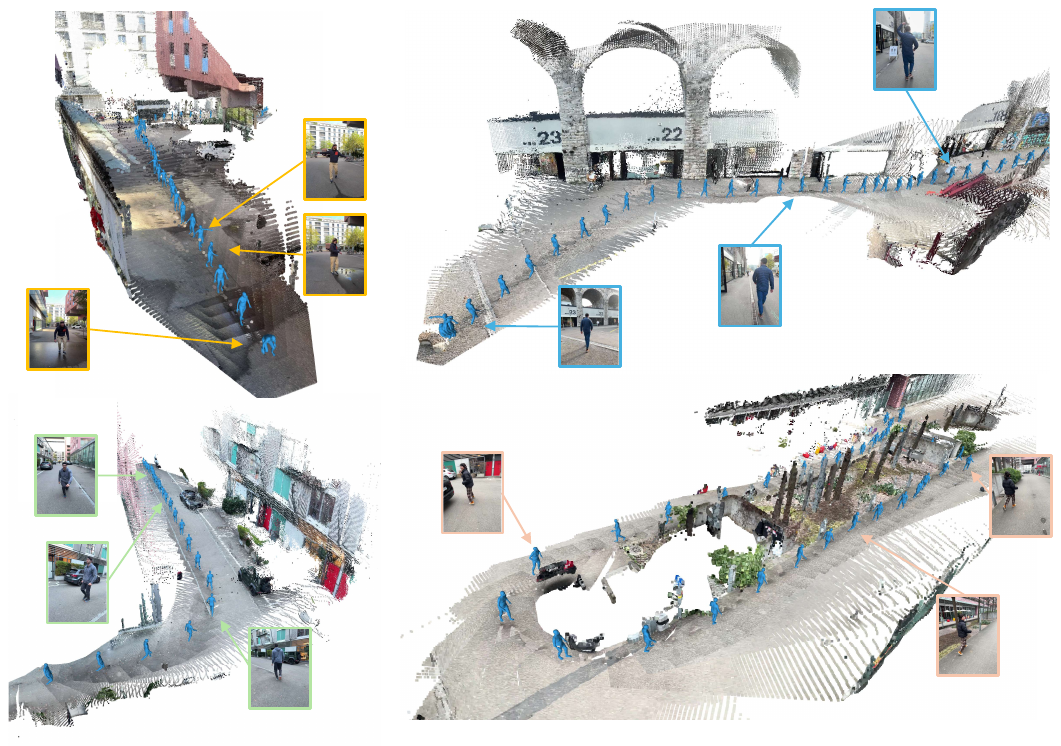}
    \vspace{-5pt}
    \caption{Quality results of proposed 4D Human~\&~Scene Reconstruction pipeline on EMDB dataset.} 
    % \vspace{-10pt}
    \label{fig:emdb}
\end{figure}

\mypara{Results.}
We present 3 variants in \cref{tab:emdb}: the proposed baseline with the original checkpoints from $\pi^3$~\cite{pi3} and VIMO~\cite{tram}, fine-tuning only VIMO, and fine-tuning both $\pi^3$ and VIMO. The results demonstrate that our approach significantly improves the accuracy of VIMO, as we provide paired high-quality real-world RGB sequences and ground truth SMPL parameters. Additionally, leveraging our high-quality RGB-D data and camera parameter pairs, $\pi^3$'s ability to predict in the world coordinate system also shows improvement. Our pipeline shows good performance on large-scale real-world videos, see \cref{fig:emdb}

\input{tables/exp_mocap_sota}
% \vspace{-10pt}

%% file: tables/exp_mocap_sota.tex
\begin{table}[htbp]
\centering
\caption{Comparison of Finetuned Models on EMDB Benchmarks}
\resizebox{0.7\linewidth}{!}{ 
\begin{tabular}{cc|ccc}
\toprule
\multicolumn{2}{c|}{Finetuned} &
\multicolumn{3}{c}{EMDB} \\ 
Pi3 & VIMO & WA-MPJPE$\downarrow$ & W-MPJPE$\downarrow$ & RTE$\downarrow$ \\ \midrule
\ding{55} & \ding{55} & 83.56 & 229.04 & 1.78 \\
\ding{55} & \ding{51} & 82.89 & 222.93 & 1.73 \\
\ding{51} & \ding{51} & \textbf{82.21} & \textbf{220.65} & \textbf{1.71} \\
\bottomrule
% \vspace{-8pt}
\end{tabular}}
\label{tab:emdb}
\end{table}

%% file: sections/subsections/exp2_skill.tex
\mypara{Motivation.}
We train several human-object interaction skills to demonstrate the physical realism of our approach and the scalability of our capture framework to new interaction skills. We aim to prove the efficiency and quality superiority of our framework over optical capture and monocular estimation methods.

\mypara{Implementation.}
Following \cite{amp, tokenhsi, sims}, we train physical character policies use goal-conditioned reinforcement learning to formulate character control as a Markov Decision Process (MDP) defined by states, actions, transition dynamics, a reward function $r$, and a discount factor $\gamma$. The reward $r_t \in \mathcal{R}$ is calculated by a style reward $r_t^{style}$~\cite{amp} and a task reward $r_t^{task}$. 
The policies are trained to maximize the expected discounted return:
$    J(\pi) = \mathbb{E}_{p(\tau | \pi)}\left[ \sum_{t=0}^{T-1}\gamma^{t} r_{t} \right],$
where $T$ is the episode length, $\gamma \in [0, 1]$ is the discount factor, and $r_t$ is the reward at time step $t$. 
 We use the widely adopted Proximal Policy Optimization (PPO) algorithm \cite{ppo} to train the control policy model.

Following \cite{tokenhsi, sims, interphysics}, we train a set of human object interaction skills in simulator~\cite{isaacgym}, including \textit{follow}, \textit{climb}, \textit{sit}, and \textit{lie}. These common interaction skills are designed to guide the character's root joint to reach specific target positions in 3D environments while maintaining physically realistic and motion divisty. We train these four common skills on 3 different input data: optical captured, which are collected from AMASS~\cite{amass} and SAMP~\cite{samp} following TokenHSI~\cite{tokenhsi}; ours, by segmenting the reconstructed motions into skill clips; monocular, by using the motion predicted by GVHMR~\cite{gvhmr} which is commonly used in humanoid reference motion prediction~\cite{hdmi, asap}, segmented with the same temporal slices as ours. We also train 2 extra interaction skills which have not been implemented in previous physics-based human object interaction papers: Prone and Support. 
We will illustrate the observation, reward designs, and the training details of each skill in Supp.Mat.

\mypara{Metrics.}
We follow \cite{samp, unihsi} that uses \emph{Success Rate} and \emph{Contact Error} as the main metrics to measure the quality of interactions quantitatively. Success Rate records the percentage of trials that humanoids successfully complete the contact within a certain threshold. We follow \cite{unihsi, interscene, interphysics} in setting the thresholds for various actions: 20cm for Sit, Follow, and Climb; 30cm for Lie and Prone; and 10cm for Support. For Support, the error is defined as the distance from the object surface center to the hand center, while also taking into account the distance between the two feet. Please see details in Supp.Mat.
We evaluate motion diversity using Average Pairwise Distance (APD)~\cite{case}, which measures the average pairwise distance between joint rotations and positions in generated samples. Higher APD values indicate greater diversity.

\mypara{Results.} We can find in \cref{tab:basic_skill}, for skills such as Follow, Climb, and Sit, the inherent difficulty is relatively low, and all three data settings achieve good results, very close to 100\%. Although the quality of our data is slightly inferior to optically captured data, we provide more variety of task completion trajectories and motion diversities, which contribute to improve task performance. To prove this, we ablate on skills trained with different data proportions. 1X and 2X indicate the ratio of the number of clips relative to the optical capture data. On the 4 common skills, we observe a general trend where increased data amount leads to improvements in success rate, contact error, and APD metrics.

\input{tables/exp_skill}
\begin{figure}[t] 
    \centering
    \begin{subfigure}[b]{0.95\linewidth} 
        \centering
    \includegraphics[width=\linewidth]{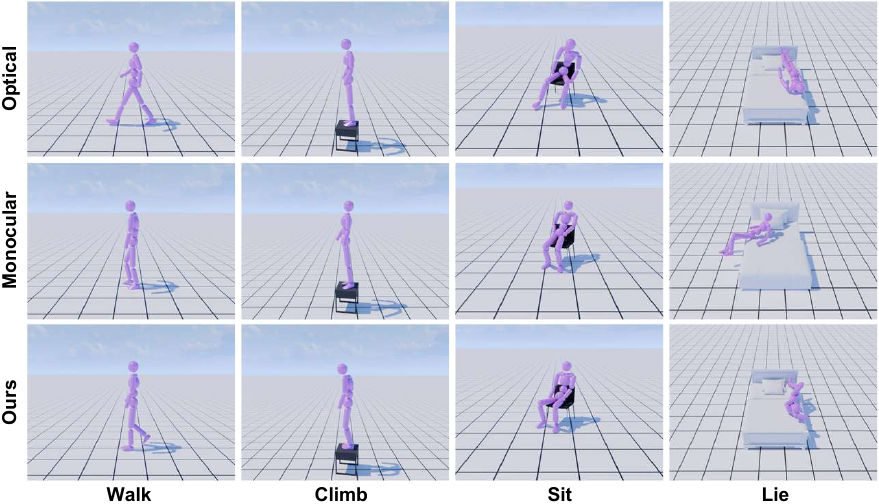} 
        % \caption{Qualitative comparison on 4 basic skills.}
         \label{fig:basic_skillsa}
    \end{subfigure}
    % \vspace{-5pt} %
    \begin{subfigure}[b]{0.95\linewidth}
    \centering
    \includegraphics[width=\linewidth]{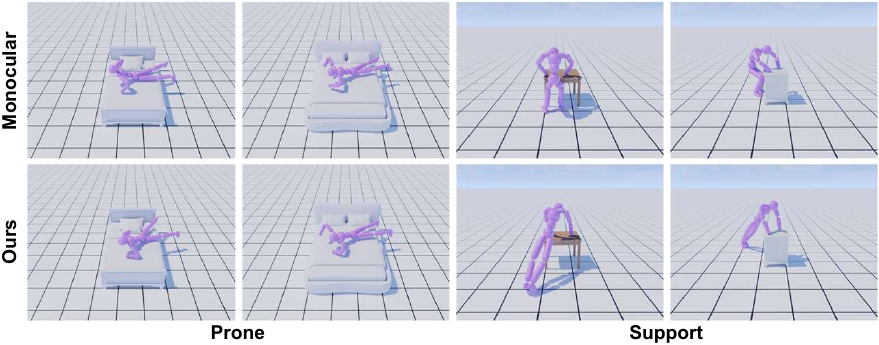}
        % \caption{Qualitative comparison on 2 additional skills.}
        \label{fig:basic_skillsb}
    \end{subfigure}
    % \vspace{-5pt}
    \caption{Qualitative comparison of 4 basic skills and 2 additional skills.}
    \label{fig:basic_skills}
\end{figure}
The success in simulating the 2 extra skills, Prone and Support, demonstrates the versatility of our data collection pipeline. First, these new skills highlight the ability of our approach to generalize to novel interaction tasks. Second, the Support skill significantly increases the level of difficulty. Unlike other tasks, where a humanoid only needs to walk or offload the full body weight onto furniture surface, Support requires the hands to bear the weight of the body while the feet remain close together, demanding much higher accuracy in reference motion generation. This experiment shows that our approach outperforms monocular estimation methods by a large margin, particularly for high-difficulty interaction skills. The success rate trained on monocular estimated motions degrades to only 20\% in \cref{tab:basic_skill}. In \cref{fig:basic_skillsb}, we can see policy trained on motion estimated from monocular models could not perform standard Support skill.

%% file: tables/exp_skill.tex
\begin{table}[t]
  \begin{center}
    \caption{Comparison of data duration, Success Rate, Contact Error, and APD for different skills among 3 data settings.}
    % \vspace{-7pt}
  \resizebox{0.98\linewidth}{!}{
    \begin{tabular}{l|C{65pt}|C{30pt}|C{60pt}|C{45pt}|C{50pt}|C{60pt}}
    \toprule[1.5pt]
    Task & Data   & Clips& Duration~(min) & Rate~(\%)~$\uparrow$ & Error (cm)~$\downarrow$  & APD~$\uparrow$\\
    \midrule
    \multirow{6}{*}{Follow} & Optical Mocap & 12 & 1.59 & \textbf{99.9} & \textbf{6.0} & \textbf{20.17 $\pm$ 0.19}\\
    
     &Ours 1X   & 12 & 1.48 & \textbf{99.9} & 6.7 &18.42 $\pm$ 0.22 \\
      &Ours 2X   & 24 &  3.06 & 99.7 &  6.8 & 18.45 $\pm$ 0.17 \\
       % &Ours 4X   & 48 & 11.1 & 99.8 & 6.3 \\
    & Ours Full   & 148 &  22.43 & 99.8 & \underline{6.2} & 19.69 $\pm$ 0.32\\
    & Monocular  & 148 &  22.43 &  98.0 & 7.2  & \underline{19.85 $\pm$ 0.39} \\
    
    \midrule

    \multirow{5}{*}{Climb}  & Optical Mocap & 7 & 0.28 & \textbf{99.9} & 2.7 & 22.03 $\pm$ 0.30  \\
        
     &Ours 1X   & 7 & 0.54 & 99.8 & \textbf{1.8}  & \textbf{22.77} $\pm$ 0.29 \\
      &Ours 2X   & 14 & 0.97 & \textbf{99.9} & \textbf{1.8} & 20.72 $\pm$ 0.30 \\
       % &Ours 4X   &   & \\
    & Ours Full     &21 & 1.54 & \textbf{99.9} & \textbf{1.8} & \underline{22.22 $\pm$ 0.27} \\
     & Monocular  & 21 & 1.54 & 99.2 & 1.8 & 21.34 $\pm$ 0.38\\
    
    \midrule
    \multirow{5}{*}{Sit}    & Optical Mocap & 20 & 4.08  &98.0 & 5.5 & \textbf{16.07 $\pm$ 0.39} \\
        
     &Ours 1X   & 20 & 2.11 &99.8& 5.4 & 14.35 $\pm$ 0.27 \\
      &Ours 2X   & 40 & 4.47 & \textbf{99.9} & \underline{5.1} & 14.46 $\pm$ 0.24\\
       % &Ours 4X   &   & \\
        & Ours Full     & 80 & 8.05 & 
        \textbf{99.9} & \textbf{4.7} & \underline{15.90} $\pm$ 0.51 \\
     & Monocular & 80 & 8.05 &98.4 & 5.7 & 15.80 $\pm$ 0.51 \\
        
    \midrule
    \multirow{6}{*}{Lie}    & Optical Mocap & 10 & 2.52    & \underline{89.0} & \textbf{17.5} & \textbf{8.76 $\pm$ 0.14}   \\
        
     &Ours 1X   & 10 & 0.99  & 85.3 & 20.2  & 7.43 $\pm$ 0.10 \\
&Ours 2X   & 20 & 2.32 &  86.3 & 19.8 & 8.27 $\pm$ 0.06 \\     
            
    & Ours Full     & 39 & 4.25   & \textbf{89.4} & \underline{18.8} & \underline{8.57 $\pm$ 0.10} \\
     & Monocular  & 39 & 4.25  & 81.2 & 21.0 &  8.14 $\pm$ 0.10\\ 
     % \midrule
     
    % \multirow{6}{*}{Carry}  & Optical Mocap & 4 & 0.3 &90.1 & 5.5 \\
        
    %  &Ours 1X   & 4 & 0.3 & 74.6 & 4.4 & \\
    %         &Ours 2X   & 8 & 0.5  & \\      
    %         &Ours 4X   & 16 & 1.1 & \\
    % & Ours Full     & 87 & 6.5    & -   \\
    %  & Monocular & 87 & 6.5  \\
    
    \midrule[1pt]
    % \multirow{2}{*}{Crawl} \\
    % & \\\midrule
    \multirow{2}{*}{Prone}  &  Ours Full & 3 & 0.26 & \textbf{75.4} & \textbf{16.5} & \textbf{17.58 $\pm$ 0.69} \\
    & Monocular & 3 & 0.26  & 71.2 & 16.5 & 16.18 $\pm$ 0.30 \\ \midrule

     \multirow{2}{*}{Support} &  Ours Full & 8 & 0.97  & \textbf{66.0} & \textbf{4.9} & \textbf{21.08 $\pm$ 0.59} \\ 
    & Monocular & 8 & 0.97 & 20.6 & 6.4 & 20.94 $\pm$ 0.48 \\
    % \midrule
    % \multirow{2}{*}{Handstand}  \\
    % \\
    % \multirow{2}{*}{Cartwheel} \\
    % & \\
    \bottomrule[1.5pt]
    \end{tabular}}
    \label{tab:basic_skill}
    % \vspace{-10pt}
  \end{center}
\end{table}

%% file: sections/subsections/exp2_track.tex
\begin{figure*}[!h]
    \centering
    % 设置占位框的宽度和高度
    \includegraphics[width=0.98\linewidth]{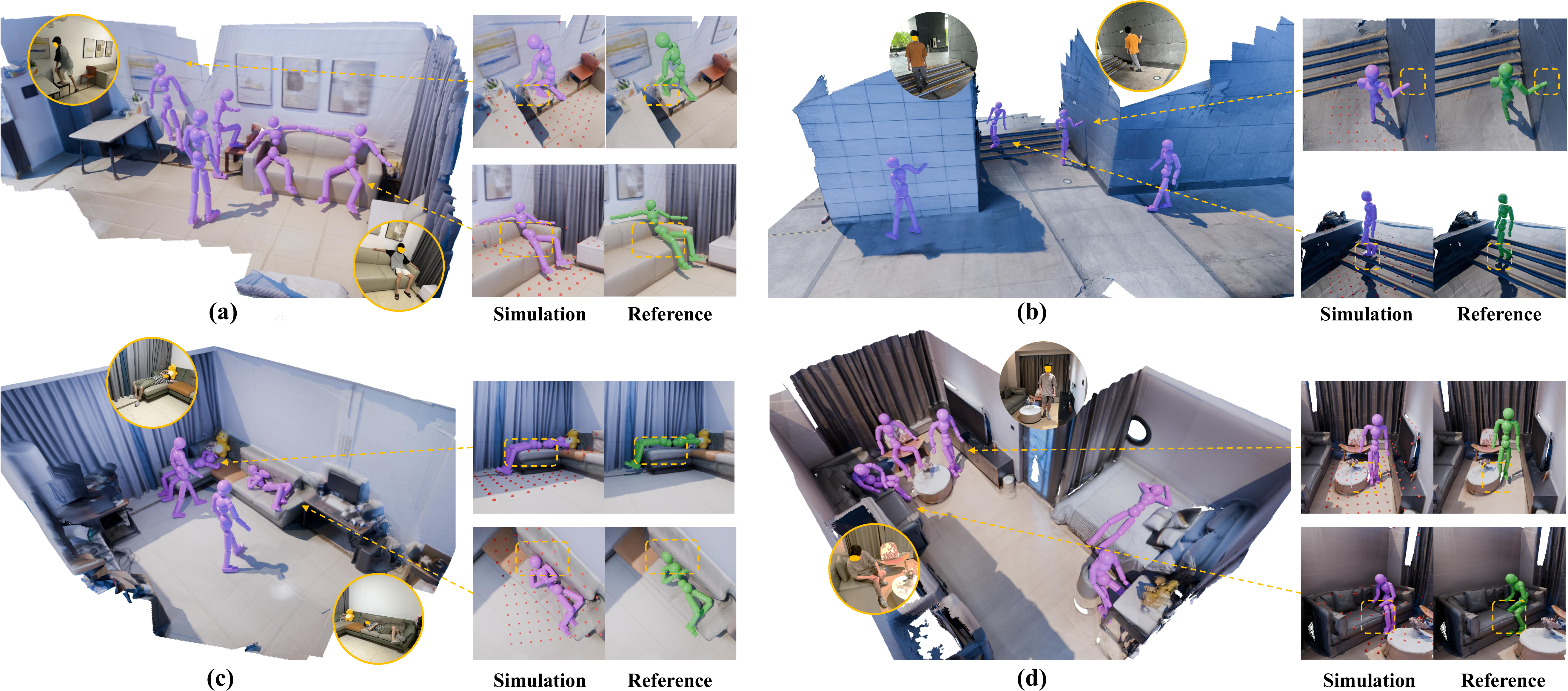}
    % \vspace{-6pt}
    \caption{We present qualitative results of scene-aware motion tracking, showing four long-term motion examples in different scenes (a, b, c, and d), including daily indoor and outdoor interactions such as walking, sitting, lying, stair climbing, and touching. Our motion tracking framework not only accurately tracks the reference motion but also ensures physical realism, resolving subtle issues, such as interpenetration and floating artifacts, present in the reference data (see zoomed-in views on the right).}
    \label{fig:track}
    % \vspace{-5pt}
\end{figure*}

\mypara{Motivation.}
Recent works~\cite{ase,pulse,ominigrasp,yao2024moconvq,tessler2024maskedmimic,tessler2025maskedmanipulator,tirinzonizero} suggest that solving complex tasks requires pre-training on large-scale human motion data via motion tracking objectives, in order to obtain reusable and generalizable skill priors. However, existing motion tracking frameworks are mainly built for human-only~\cite{phc} or single-object interaction~\cite{xu2025intermimic} scenarios, primarily because current public datasets are concentrated in these settings. We argue that motion tracking pre-training on diverse 3D scenes is equally important, as it also provides rich priors—such as navigation, interaction, and long-horizon task execution. In this work, we mitigate this gap by: 1) proposing a scene-aware motion tracking framework, and 2) supporting it with high-fidelity paired 3D human-scene data captured by our EmbodMocap system.

\mypara{Implementation.} We extend MimicKit~\cite{MimicKitPeng2025} by incorporating the height map into the observation space to achieve scene-aware tracking (details in the Supp. Mat.). For training, we use four 3D scenes, each containing several minutes of motion clips, and train one policy per scene to track all the motion clips in that scene.

\mypara{Metrics.} Policies are evaluated using a success rate metric: an episode is initialized from a random frame and run for 10s, and is considered successful if tracking exceeds 8s. For each scene, 3,072 episodes are used to compute average success, failure rates, and episode length statistics.
\input{tables/exp_track}

\mypara{Results.} The quantitative results in Tab.~\ref{tab:scene_track} demonstrate that our data is simulation-ready, enabling the training of scene-aware tracking policies with high success rates. The qualitative results, shown in Fig.~\ref{fig:track}, further illustrate that the policies not only successfully track the motions but also adapt to subtle imperfections present in the data.

%% file: tables/exp_track.tex
\begin{table}[t]
  \begin{center}
    \caption{Quantitative evaluation of scene-aware motion tracking and dataset statistics across four 3D scenes.}
    % \vspace{-7pt}
  \resizebox{0.9\linewidth}{!}{
    \begin{tabular}{c|C{30pt}|C{60pt}|C{50pt}|C{50pt}|C{60pt}}
    \toprule[1.5pt]
    
    Scene & Clips & Duration~(min) & Status & Rate~(\%)  & Eps. Len.~(s)\\
    \midrule

    \multirow{2}{*}{a}  
     & \multirow{2}{*}{14} & \multirow{2}{*}{12.31} & Succ. & 87.2 & 9.97 $\pm$ 0.21  \\
     & & & Fail. & 12.8  & 3.94 $\pm$ 2.10 \\
    \midrule

    \multirow{2}{*}{b}    
     & \multirow{2}{*}{6} & \multirow{2}{*}{3.62}    & Succ. & 96.7 & 9.99 $\pm$ 0.12   \\
     & &  & Fail. & 3.3  & 4.16 $\pm$ 2.38 \\
     \midrule

     \multirow{2}{*}{c}    
     & \multirow{2}{*}{12} & \multirow{2}{*}{7.87}   & Succ. & 95.9 & 9.98 $\pm$ 0.17 \\
     &  &  & Fail. & 4.1 & 5.43 $\pm$ 2.18 \\
    
    \midrule

     \multirow{2}{*}{d} 
     & \multirow{2}{*}{7}  & \multirow{2}{*}{5.06} & Succ. & 90.4 & 9.96 $\pm$ 0.21\\
     & & & Fail. & 9.6 & 4.44 $\pm$ 1.92 \\

    \bottomrule[1.5pt]
    \end{tabular}}
    \label{tab:scene_track}
    \vspace{-5pt}
  \end{center}
\end{table}

%% file: sections/subsections/exp3_robo.tex
\begin{figure}[t]
    \centering
    % 设置占位框的宽度和高度
    \includegraphics[width=0.95\linewidth]{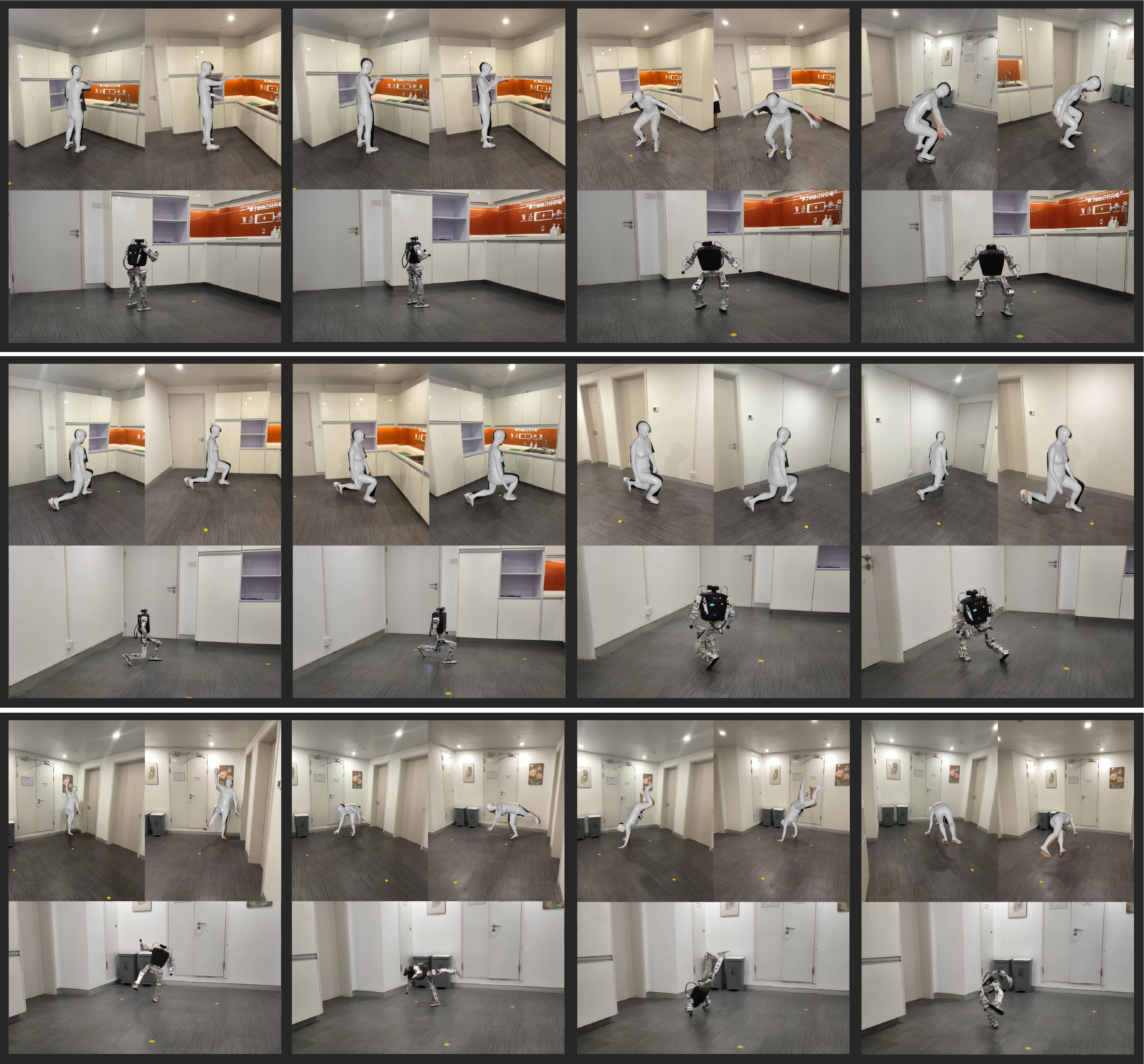}
    \caption{A real-world humanoid robot imitating human motions depicted in videos.}
    % \vspace{-15pt}
    \label{fig:robot}
\end{figure}

\mypara{Motivation.} Learning from human videos~\cite{qiu2025humanoid,hdmi,videomimic} has emerged as a crucial paradigm for humanoid robots to learn motor skills at scale. In this section, we demonstrate how EmbodMocap contributes to this paradigm by enabling accurate reconstruction of humans and their interacting 3D environments from videos, while preserving accurate contact information.

\mypara{Implementation.} We capture videos of humans performing ground-contact-rich motions, including locomotion and challenging cartwheels that require precise hand-ground contact. EmbodMocap is then used for real-to-sim reconstruction. The produced motions are used to train a single tracking policy via sim-to-real RL with domain randomization using BeyondMimic~\cite{liao2025beyondmimic}.

\mypara{Results.} We deploy the policy on a real-world High Torque Hi humanoid robot with 21 joint DoF and a height of 80cm. As shown in Fig.~\ref{fig:robot}, the robot successfully replicates human motions from videos, demonstrating that EmbodMocap produces data of sufficient quality for humanoid robot control.

%% file: sections/5_conclusion.tex
\section{Conclusion}
We propose EmbodMocap, a portable and affordable framework for capturing high-quality 4D human \& scene data using only two iPhones. Our method enables scalable, metrically accurate reconstruction of human motion and scenes mesh in diverse real-world environments. We directly compare in optical capture studios, and prove the superiority in solving body occlusion and sequence coordinate alignment of our dual view designing. Through downstream applications in monocular human-scene reconstruction, physics-based character animation, and humanoid robot motion control, we demonstrate the effectiveness and scalability of our approach. By lowering the barrier for embodied AI research, EmbodMocap opens new opportunities for real-world applications.

\section{Limitations and Future Work.} 
Our data collection pipeline encounters limitations in specific scenarios. For example, it fails to record depth when the distance exceeds the range of the iPhone LiDAR sensor (approximately 5 meters). Additionally, it struggles with scenes dominated by moving objects, which degrade the results of the SLAM SDK~\cite{spectacularai}. Extremely bright lighting conditions can also cause COLMAP failures, leading to incorrect registration. Future work could integrate more robust structure-from-motion tools, such as H-Loc~\cite{hloc}, to improve reliability. Moreover, incorporating automatic synchronization APPs on iPhone could further reduce human effort.

%% file: sections/6_acknowledge.tex
\section{Acknowledge}
This work was partially funded by the Innovation and Technology Commission of the HKSAR Government under the ITSP-Platform grants (Ref: ITS/335/23FP, ITS/469/24FP). 

We sincerely thank Mr. Xiaohan Ye, Mr. Rui Xu and Mr. Kaiyuan Zheng for volunteering as actors during data collection.

%% file: sections/X_supp.tex
\clearpage
\setcounter{page}{1}
\maketitlesupplementary

% \section{Core Contribution Credit}
% \mypara{Capture System (Section 3)}
% \begin{itemize}[left=1em]
%     \item Wenjia Wang: Designing, Main Conductor.
%     \item Huaijin Pi: Discussion on Designing
%     \item Xuqian Ren: Assistance on COLMAP and SAI

% \end{itemize}

% \mypara{Experiment of 4DHuman \& Scene (Section 3.1)}
% \begin{itemize}[left=1em]
%     \item Wenjia Wang: Designing, Main Conductor.
%     \item Huaijin Pi: Discussion \& Assistance
% \end{itemize}

% \mypara{Experiment of HOI Skills (Section 4.2.1)}
% \begin{itemize}[left=1em]
%     \item Wenjia Wang: Designing, Main Conductor.
% \end{itemize}

% \mypara{Experiment of Scene-Aware Imitation (Section 4.2.2)}
% \begin{itemize}[left=1em]
%     \item Liang Pan: Designing, Main Conductor.
%     \item Wenjia Wang: Discussion \& Assistance
% \end{itemize}

% \mypara{Experiment of Real World Humanoid (Section 4.3)}
% \begin{itemize}[left=1em]
%     \item Liang Pan: Designing, Main Conductor.
%     \item Wenjia Wang: Discussion \& Assistance
% \end{itemize}

% \mypara{Writing}
% \begin{itemize}[left=1em]
%     \item Wenjia Wang
%     \item Huaijin Pi
%     \item Liang Pan
%     \item Yifan Wu
% \end{itemize}

% \mypara{Visualization}
% \begin{itemize}[left=1em]
%     \item Wenjia Wang
%     \item Yuke Lou
%     \item Yifan Wu
%     \item Liang Pan
% \end{itemize}

% \mypara{Dataset Capture}
% \begin{itemize}[left=1em]
%     \item Wenjia Wang
%     \item Yuke Lou
%     \item Yifan Wu
%     \item Acknowledge: Mr. Xiaohan Ye and Mr. Rui Xu
% \end{itemize}

\section{More Details of EmbodMocap}
\subsection{Capture technique}
The primary capture technique involves two photographers, each holding an iPhone in a vertical orientation. The photographers are required to maintain a certain angle relative to each other while following the performer. To achieve optimal triangulation during post-processing, the angle between the two cameras should ideally fall within the range of 60 to 120 degrees.

This configuration not only enhances the accuracy of triangulation but also ensures the capture of the performer from multiple perspectives, providing diverse viewpoint information for keypoint detection. Additionally, the photographers should aim to keep the cameras in motion to dynamically adjust their positions and minimize occlusion caused by objects in the environment.
\begin{figure}[h]
    \centering
    \includegraphics[width=0.9\linewidth]{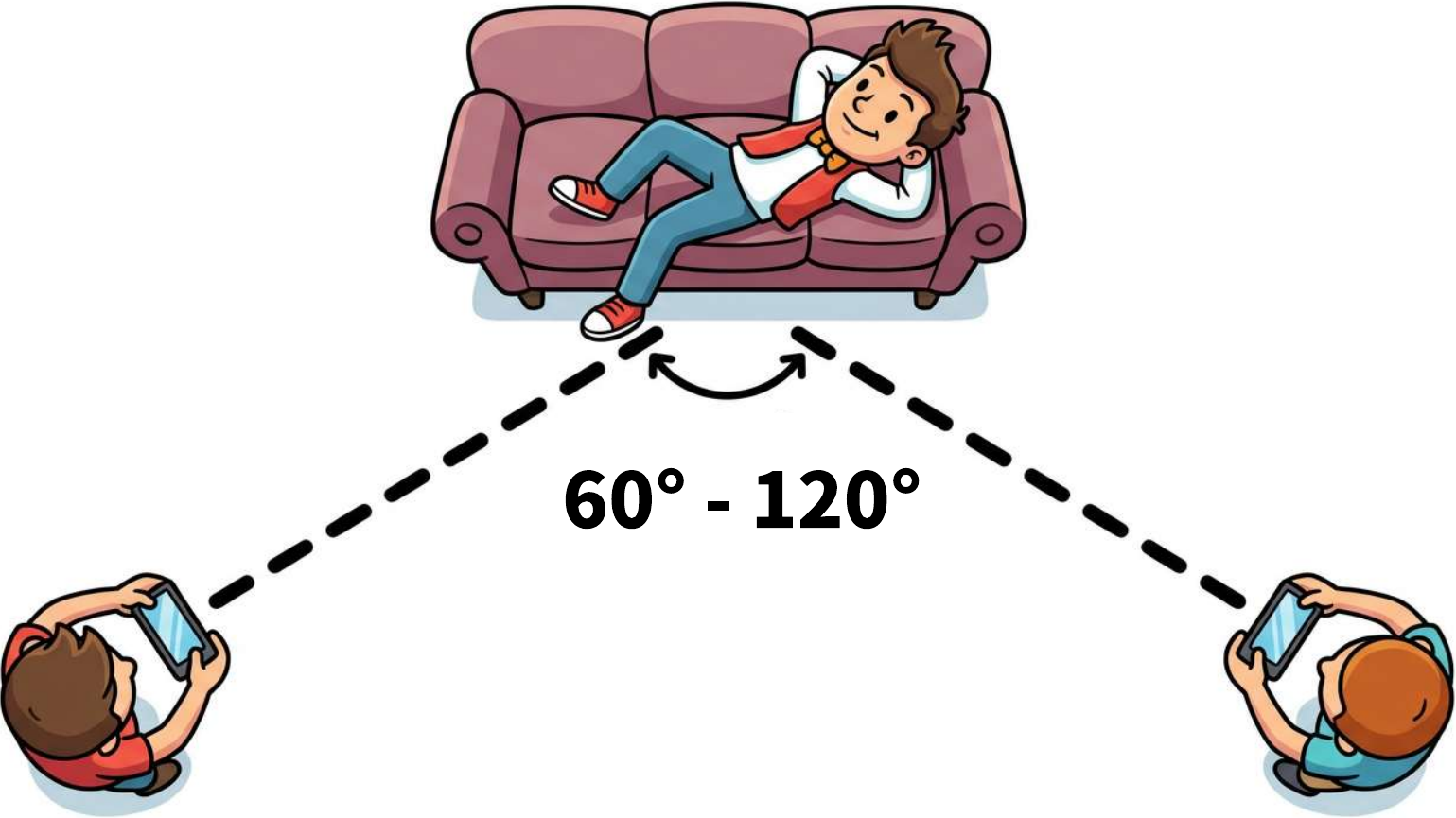}
    \caption{Capture technique.}
    \label{fig:placeholder}
\end{figure}

\subsection{Human Labor Analysis}
\label{sec:human_labor}
\mypara{Temporal Synchronization.} 
This step only needs the operator to identify and input the frame indices where the laser pointer's spot disappears into a $\texttt{.xlsx}$ file. Typically, this process takes only about 1 minute per sequence.

\mypara{Skill Segmentation.}
Skill segmentation is only required when training physical interaction skills. The operator annotates each skill's category, start, and end times based on the video, typically taking 0.5 to 2 minutes per sequence.

\mypara{Contact Label \& Optimization.}
In the main text, we mention that the alignment between our sequence and the scene coordinate system relies on photometric (COLMAP, pixel tracking) and geometric constraints (chamfer distance). However, this can sometimes result in alignment errors of a few centimeters, primarily due to depth inaccuracies in COLMAP's sparse keypoints and depth errors from the iPhone sensor. To address this issue, we propose an optional post-processing solution. During data capture, we place markers in the scene and instruct the performer to begin walking from a designated marker and stop on another at the end of the sequence, standing still on the same marker. Annotating contact frame indices costs 1-2 minutes for each sequence. These markers serve as fixed reference points for alignment. In post-processing, we observe the corresponding marker positions on the reconstructed mesh and record their 3D coordinates, along with the frame indices where the performer stands on the markers. Using this information, we optimize a rigid transformation to align the center of the performer’s feet at the specified frame indices to the 3D coordinates of the markers. 

Since SpectacularAI could generate Z-up metric-scaled camera matrices, we define the rigid transformation in the xy-plane, defined by a rotation angle $\phi_c$ about the z-axis and a translation $\bm{T}_{c}$. This can be represented by a homogeneous transformation matrix $\bm{M}$:
{
\begin{equation}
M = \begin{bmatrix} \bm{R}(\phi_c) & \bm{T}_c \\ \bm{0} & 1 \end{bmatrix} = \begin{bmatrix} \cos(\phi_c) & -\sin(\phi_c) & 0 & t_x \\ \sin(\phi_c) & \cos(\phi_c) & 0 & t_y \\ 0 & 0 & 1 & t_z  \\ 0 & 0 & 0 & 1 \end{bmatrix}
\end{equation}
}
This matrix transform the center of lowest point on both feet to match the annotate marker. To robustly solve for the transformation parameters, we employ a gradient descent optimization, constrained by a minimizing a contact loss to match the contact marker:
{
\begin{align}
\mathcal{L}_{\text{contact}} &= \frac{1}{N_c} \sum_{i \in \mathcal{C}} \left( \min_{z}(\mathcal{V}^{(i)}) - c_z^{(i)} \right)^2 
\end{align}
} 
For SMPL parameters, the global orientation is updated as $\bm{\theta}'^g = \bm{R}_c \bm{\theta}^g$. For translation, the pelvis's world position is transformed as $\bm{P}'_{w} = \bm{R}_c \bm{P}_{w} + \bm{T}_c$. Re-evaluating the SMPL model with $\bm{\theta}'^g$ gives the local pelvis offset $\bm{P}'_l$, and the updated translation is $\bm{\gamma}' = \bm{P}'_w - \bm{P}'_l$.

The updated camera rotation and translation are computed as $\bm{R}_v' = \bm{R}_v \bm{R}c^T$ and $\bm{T}_v' = \bm{T}_v - \bm{R}_v \bm{R}c^T \bm{T}_c$, ensuring alignment and consistency of the scene representation.
\begin{figure}
    \centering
    \includegraphics[width=0.95\linewidth]{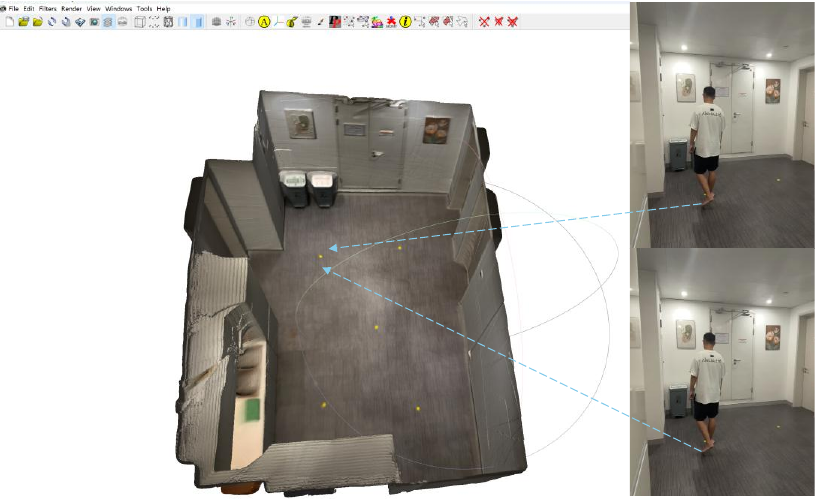}
    \caption{An example in finding the contact marker in software (e.g., Meshlab) and corresponding keyframe index(the frames selected here are just for demo).}
    \label{fig:placeholder}
\end{figure}

\section{More Details of Monocular Human-Scene Reconstruction Pipeline}

% \subsection{Some Pipeline Details}
Our monocular reconstruction baseline is a modular pipeline for reconstructing 3D human pose and scene geometry from monocular video, combining two independent modules: $\pi^3$ for camera trajectory prediction and scene point cloud reconstruction, and VIMO for SMPL-based human pose estimation. To process long video sequences, $\pi^3$ divides frames into overlapping chunks, where each chunk independently predicts camera poses $\bm{T}_v \in \mathbb{R}^{T \times 4 \times 4}$ and local point clouds $\bm{P}_{\text{local}} \in \mathbb{R}^{T \times H \times W \times 3}$. To align these chunks into a global coordinate system, Procrustes analysis is applied to the overlapping regions of adjacent chunks. Given two point clouds $\bm{X}, \bm{Y} \in \mathbb{R}^{N \times 3}$, the alignment minimizes the error:
{
\begin{equation}
\min_{s, \bm{R}, \bm{t}} \|\bm{Y} - (s \bm{R} \bm{X} + \bm{t})\|_F^2,
\end{equation}
}
where $s$ is the scale, $\bm{R}$ is the rotation matrix, and $\bm{t}$ is the translation vector. Using SVD, the optimal alignment parameters are computed as: 
{
\begin{equation}
\bm{R} = \bm{V} \bm{S} \bm{U}^\top, \quad s = \frac{\text{trace}(\bm{Y}_c^\top \bm{R} \bm{X}_c)}{\text{trace}(\bm{X}_c^\top \bm{X}_c)}, \quad \bm{t} = \bar{\bm{Y}} - s \bm{R} \bar{\bm{X}},
\end{equation}
}
where $\bm{X}_c, \bm{Y}_c$ are the centered point clouds, and $\bm{V}, \bm{U}$ are derived from the SVD of the covariance matrix $\bm{H} = \bm{X}_c^\top \bm{Y}_c$. After chunk alignment, VIMO predicts SMPL parameters $(\bm{\theta}, \bm{\gamma}, \bm{\beta})$, where $\bm{\theta} \in \mathbb{R}^{T \times 72}$ represents joint rotations, $\bm{\gamma} \in \mathbb{R}^{T \times 3}$ is the root translation, and $\bm{\beta} \in \mathbb{R}^{10}$ defines body shape. Using a weak perspective camera model, SMPL vertices are projected onto the image plane as:
{
\begin{equation}
\bm{x}_{\text{img}} = s \bm{x}_v + \bm{t}
\end{equation}
}
where $s$ is the scaling factor proportional to $1/z$. To resolve scale ambiguity, the pipeline estimates a metric scale by matching the predicted depths of SMPL vertices $z_{\text{SMPL}}$ (in meters) with the depths of Pi3’s point cloud $z_{\text{Pi3}}$ (in arbitrary units) on some sampled points. The scale factor is computed as:
{
\begin{equation}
s = \text{median}\left(\frac{z_{\pi^3}}{z_{\text{SMPL}}}\right),
\end{equation}
}

The point clouds and SMPL global orientation and translation are transformed to the world coordinate system with $\bm{R}, \bm{t}$ following the same formula as \cref{sec:human_labor}.

% \subsection{More Results}

\section{More Details of Human-Object Interaction Skills}

\subsection{Follow Skill}

\mypara{Definition.} The path following task requires the simulated character to move along a predefined 2D trajectory. A trajectory is represented as $\tau = \{ x_{0.1}^\tau, x_{0.2}^\tau, \dots, x_{T-0.1}^\tau, x_T^\tau \}$, where $x_{0.1}^\tau$ denotes a 2D waypoint at simulation time $0.1s$, and $T$ is the episode length. For this task, $T$ is set to $10s$. The character is expected to follow the trajectory $\tau$ as accurately as possible.

\mypara{Task Observation.} At each simulation time step $t$, the character observes $10$ future waypoints sampled over the next $1.0s$: $\{ x_t^\tau, x_{t+0.1}^\tau, \dots, x_{t+0.8}^\tau, x_{t+0.9}^\tau \}$. These waypoints are sampled at intervals of $0.1s$ using linear interpolation from the trajectory $\tau$. The 2D coordinates of these waypoints form the task observation $g_t^f \in \mathbb{R}^{2 \times 10}$.

\mypara{Task Reward.} The reward for this task, $r_t^f$, is computed based on the distance between the character's current 2D root position, $x_t^{\text{root\_2d}}$, and the target waypoint, $x_t^\tau$. The reward is defined as:
\begin{equation}
    r_t^f = \exp\big(-2.0 \| x_t^{\text{root\_2d}} - x_t^\tau \|^2 \big).
\end{equation}

\subsection{Sit Skill}

\mypara{Definition.} The sitting task requires the character to position its root joint at a target 3D sitting location on an object surface. The target position is defined as $10$ cm above the center of the top surface of the chair seat.

\mypara{Task Observation.} The observation $g_t^s \in \mathbb{R}^{38}$ includes the 3D target sitting position $\in \mathbb{R}^3$, the 3D root position $\in \mathbb{R}^3$, the root rotation $\in \mathbb{R}^6$, the 2D front-facing direction $\in \mathbb{R}^2$, and the positions of eight corner points of the object's bounding box $\in \mathbb{R}^{3 \times 8}$.

\mypara{Task Reward.} The sitting task reward $r_t^s$ encourages the character to minimize the distance between its 3D root position, $x_t^{\text{root}}$, and the target sitting position, $x_t^{\text{tar}}$. It is defined as:
\begin{equation}
    r_t^s = 
    \begin{cases} 
    0.7 \, r_t^{\text{near}} + 0.3 \, r_t^{\text{far}}, & \| x_t^{\text{obj\_2d}} - x_t^{\text{root\_2d}} \| > 0.5, \\
    0.7 \, r_t^{\text{near}} + 0.3, & \text{otherwise},
    \end{cases}
\end{equation}
where $r_t^{\text{far}}$ and $r_t^{\text{near}}$ are defined as:
\begin{equation}
    r_t^{\text{far}} = \exp\big(-2.0 \| 1.5 - d_t^* \cdot \dot{x}_t^{\text{root\_2d}} \|^2 \big),
\end{equation}
\begin{equation}
    r_t^{\text{near}} = \exp\big(-10.0 \| x_t^{\text{tar}} - x_t^{\text{root}} \|^2 \big).
\end{equation}
Here, $x_t^{\text{obj\_2d}}$ is the 2D position of the object's root, $\dot{x}_t^{\text{root\_2d}}$ is the 2D linear velocity of the character's root, and $d_t^*$ is a horizontal unit vector pointing from $x_t^{\text{root\_2d}}$ to $x_t^{\text{obj\_2d}}$.

\subsection{Climb Skill}

\mypara{Definition.} The climbing task requires the character to place its root joint at a target 3D climbing position on a given object. The target position is set $94$ cm above the center of the top surface of the object.

\mypara{Task Observation.} The observation $g_t^m \in \mathbb{R}^{27}$ includes the 3D target root position $\in \mathbb{R}^3$ and the 3D coordinates of eight corner points of the object's bounding box $\in \mathbb{R}^{3 \times 8}$.

\mypara{Task Reward.} The climbing task reward $r_t^m$ minimizes the 3D distance between the character's root, $x_t^{\text{root}}$, and the target location, $x_t^{\text{tar}}$. The reward is defined as:
\begin{equation}
    r_t^m = 
    \begin{cases} 
    0.5 \, r_t^{\text{near}} + 0.2 \, r_t^{\text{far}}, & \| x_t^{\text{obj\_2d}} - x_t^{\text{root\_2d}} \| > 0.7, \\
    0.5 \, r_t^{\text{near}} + 0.2 + 0.3 \, r_t^{\text{foot}}, & \text{otherwise},
    \end{cases}
\end{equation}
where $r_t^{\text{near}}$, $r_t^{\text{far}}$, and $r_t^{\text{foot}}$ are defined as:
\begin{equation}
    r_t^{\text{near}} = \exp\big(-10.0 \| x_t^{\text{tar}} - x_t^{\text{root}} \|^2 \big),
\end{equation}
\begin{equation}
    r_t^{\text{far}} = \exp\big(-2.0 \| 1.5 - d_t^* \cdot \dot{x}_t^{\text{root\_2d}} \|^2 \big),
\end{equation}
\begin{equation}
    r_t^{\text{foot}} = \exp\big(-50.0 \| (x_t^{\text{tar\_h}} - 0.94) - x_t^{\text{foot\_h}} \|^2 \big).
\end{equation}
Here, $x_t^{\text{tar\_h}}$ is the height of the target root position, $(x_t^{\text{tar\_h}} - 0.94)$ represents the height of the top surface of the target object in world coordinates, and $x_t^{\text{foot\_h}}$ is the mean height of the character's feet. The reward $r_t^{\text{foot}}$ encourages the character to lift its feet and is crucial for successful climbing.

\subsection{Lie Skill}

\mypara{Definition.} The lying task requires the character to position its root joint at a target 3D lying position on an object, typically centered on the object's surface. The character must first approach a designated standing point before transitioning into the lying position.

\mypara{Task Observation.} The observation $g_t^l \in \mathbb{R}^{38}$ includes the 3D target lying position $\in \mathbb{R}^3$, the 3D root position $\in \mathbb{R}^3$, the root rotation $\in \mathbb{R}^6$, the 2D front-facing direction $\in \mathbb{R}^2$, and the positions of eight corner points of the object's bounding box $\in \mathbb{R}^{3 \times 8}$. It also includes the chosen standing point $\in \mathbb{R}^3$.

\mypara{Task Reward.} The lying reward $r_t^l$ combines rewards for approaching the standing point and accurately lying down:
\begin{equation}
    r_t^l = 
    \begin{cases} 
    0.6 \, r_t^{\text{near}} + 0.4 \, r_t^{\text{far}}, & \| x_t^{\text{root}} - x_t^{\text{tar}} \| > 1.5, \\
    r_t^{\text{near}}, & \text{otherwise}.
    \end{cases}
\end{equation}

The far reward encourages approaching the standing point:
\begin{equation}
    r_t^{\text{far}} = 0.5 \, r_t^{\text{walk}} + 0.2 \, r_t^{\text{vel}} + 0.2 \, r_t^{\text{facing}} + 0.1 \, r_t^{\text{stand}},
\end{equation}
where $r_t^{\text{walk}}$ rewards walking toward the standing point, $r_t^{\text{vel}}$ aligns velocity, $r_t^{\text{facing}}$ ensures proper facing direction, and $r_t^{\text{stand}}$ rewards correct height.

The near reward focuses on lying accuracy:
\begin{equation}
    r_t^{\text{near}} = 0.5 \, r_t^{\text{pos}} + 0.3 \, r_t^{\text{head}} + 0.2 \, r_t^{\text{alignment}},
\end{equation}
where $r_t^{\text{pos}}$ minimizes the distance to the target, $r_t^{\text{head}}$ aligns head height, and $r_t^{\text{alignment}}$ rewards proper body alignment.

\subsection{Prone Skill}

\mypara{Definition.} The prone task requires the character to position its root joint at a designated 3D prone position on an object, typically centered on the object's surface. Unlike the lying task, the character must face downward while maintaining alignment with the target surface.

\mypara{Task Observation.} The observation $g_t^p \in \mathbb{R}^{35}$ includes the 3D target prone position $\in \mathbb{R}^3$, the 3D root position $\in \mathbb{R}^3$, the root rotation $\in \mathbb{R}^6$, the 2D front-facing direction $\in \mathbb{R}^2$, and the positions of eight corner points of the object's bounding box $\in \mathbb{R}^{3 \times 8}$. These observations help guide the approach and ensure the correct orientation for prone positioning.

\mypara{Task Reward.} The prone reward $r_t^p$ encourages the character to transition smoothly from moving to a prone position while maintaining proper alignment and facing downward. The reward is defined as:
\begin{equation}
    r_t^p = 
    \begin{cases} 
    0.7 \, r_t^{\text{near}} + 0.3 \, r_t^{\text{far}}, & \| x_t^{\text{root}} - x_t^{\text{tar}} \| > 1.5, \\
    r_t^{\text{near}}, & \text{otherwise}.
    \end{cases}
\end{equation}

The far reward encourages approaching the target prone position:
\begin{equation}
    r_t^{\text{far}} = 0.5 \, r_t^{\text{walk}} + 0.2 \, r_t^{\text{vel}} + 0.2 \, r_t^{\text{facing}} + 0.1 \, r_t^{\text{height}},
\end{equation}
where $r_t^{\text{walk}}$ rewards moving toward the prone position, $r_t^{\text{vel}}$ aligns velocity with the direction of motion, $r_t^{\text{facing}}$ ensures proper facing direction, and $r_t^{\text{height}}$ encourages maintaining an appropriate height during approach.

The near reward focuses on prone accuracy:
\begin{equation}
    r_t^{\text{near}} = 0.6 \, r_t^{\text{pos}} + 0.2 \, r_t^{\text{alignment}} + 0.2 \, r_t^{\text{face\_down}},
\end{equation}
where $r_t^{\text{pos}}$ minimizes the distance to the prone target, $r_t^{\text{alignment}}$ ensures proper body alignment with the surface, and $r_t^{\text{face\_down}}$ rewards the character for maintaining a face-down orientation.

\subsection{Support Skill}

\mypara{Definition.} The support task encourages the character to approach a target object and maintain stable interaction by placing its hands on the top surface while keeping stable foot placement and proper posture.

\mypara{Task Observation.} The task observation $g_t^m \in \mathbb{R}^{27}$ consists of the 3D target position of the object’s top surface center ($x_t^o, z_t^o \in \mathbb{R}^3$) and the 3D coordinates of the eight corner points of the object’s bounding box ($b_t \in \mathbb{R}^{3 \times 8}$).

\mypara{Task Reward.} The total reward $r_t^m$ is defined as:
\begin{align}
    r_t^m &= 
    \begin{cases} 
    0.4 r_t^f + 0.6 r_t^s, & \| x_t^o - x_t^r \| > 1.5, \\
    r_t^s, & \text{otherwise},
    \end{cases} \label{eq:main_reward} \\
    r_t^f &= 0.5 \exp\big(-0.5 \| x_t^o - x_t^r \|^2 \big) \\&+ 0.5 \exp\big(-2.0 \| 1.5 - d_t^* \cdot \dot{x}_t^r \|^2 \big), \\
    r_t^s &= 0.3 r_t^h + 0.2 r_t^g + 0.15 r_t^t + 0.2 r_t^o + 0.15 r_t^z,
\end{align}
where $r_t^f$ encourages the character to approach the object, and $r_t^s$ combines five components for stable interaction:
\begin{align}
    r_t^h &= 0.6 \exp\big(-20 \| z_t^h - z_t^o \|^2 \big) 
    \\ &+ 0.4 \exp\big(-5 \| x_t^{h2} - x_t^o \|^2 \big), \\
    r_t^g &= \exp\big(-50 \| z_t^f - z_g \|^2 \big), \\
    r_t^t &= \exp\big(-10 \| x_t^{fr} - x_t^{fl} \|^2 \big), \\
    r_t^o &= \exp\big(-2 \| 1.0 - (-u_t^b) \|^2 \big), \\
    r_t^z &= \exp\big(-10 \| z_t^r - z_t^o \|^2 \big).
\end{align}

Here, $x_t^o$ and $x_t^r$ denote the 2D positions of the object and the character’s root, while $z_t^o$ and $z_t^r$ are their respective heights. $x_t^{h2}$ and $z_t^h$ represent the 2D position and height of the hands. Similarly, $x_t^{fr}$, $x_t^{fl}$, and $z_t^f$ refer to the 2D positions and height of the feet, $z_g$ is the ground height, and $-u_t^b$ is the vertical component of the body’s up direction.

\begin{figure}[h!]
    \centering
    \begin{subfigure}[b]{0.23\textwidth} 
        \centering
        \includegraphics[width=\textwidth]{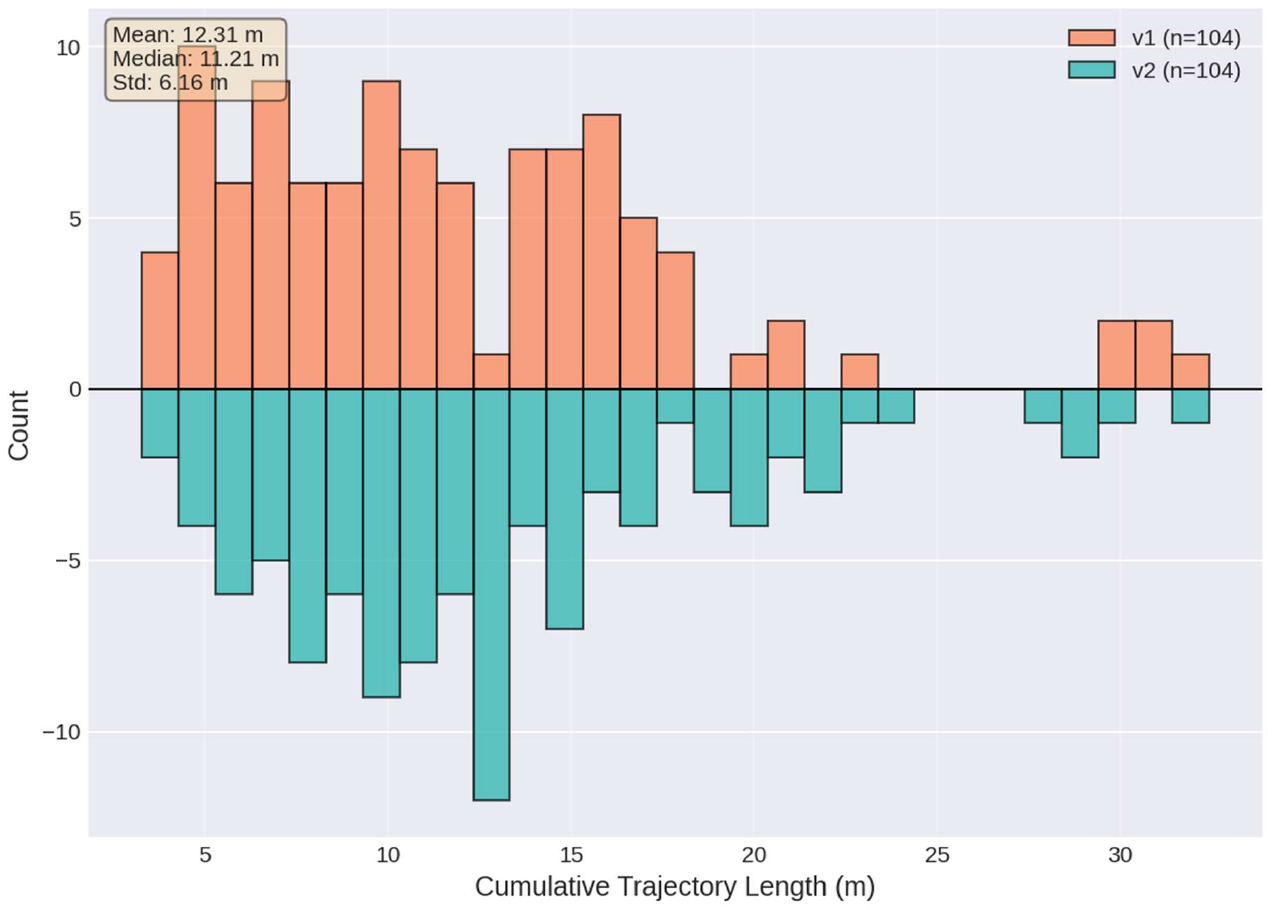} 
        \caption{Camera Trajectory Length Distribution.}
        \label{fig:stat1}
    \end{subfigure}
    \begin{subfigure}[b]{0.23\textwidth}
        \centering
        \includegraphics[width=\textwidth]{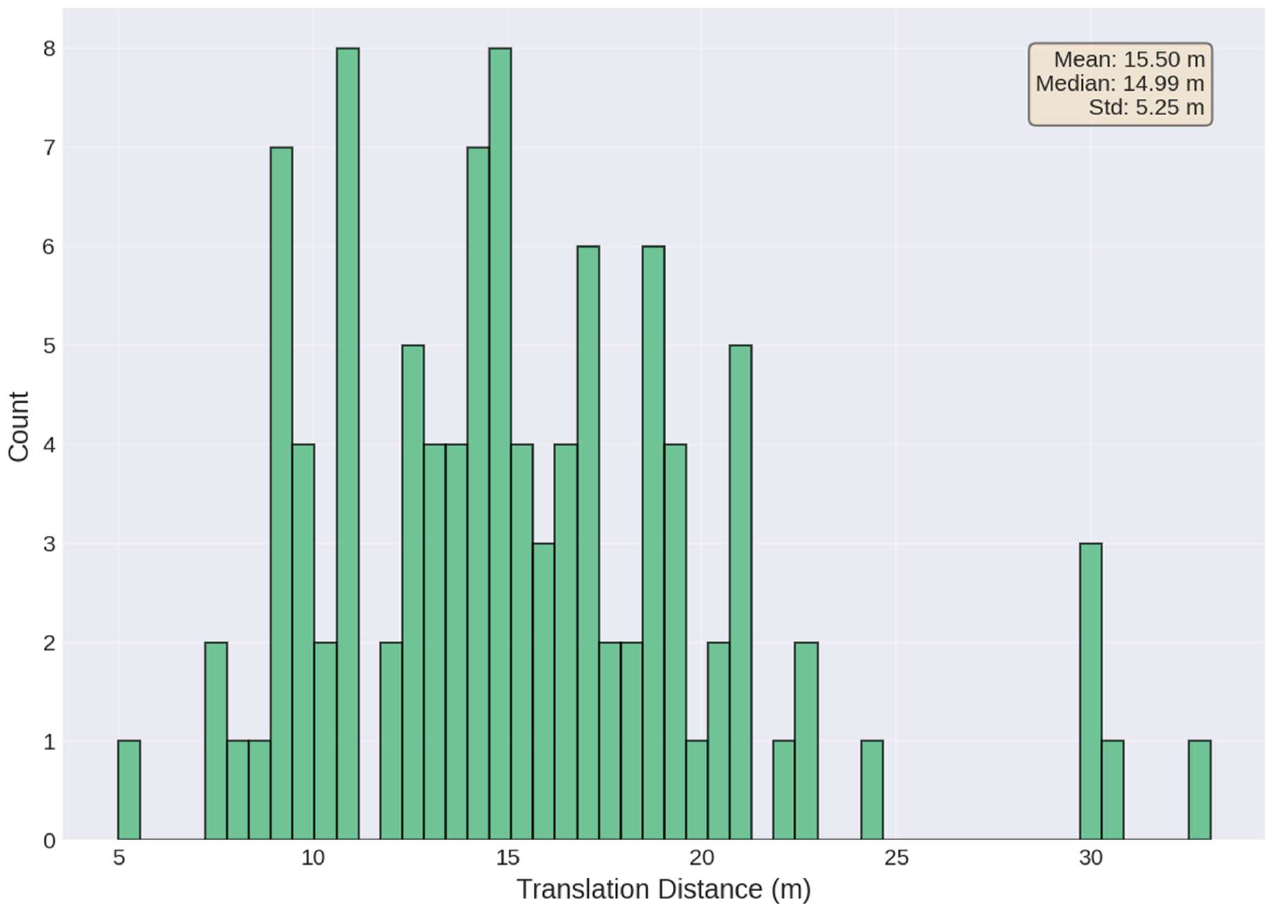} 
        \caption{Human Trajectory Length Distribution.}
        \label{fig:stat2}
    \end{subfigure}
    \begin{subfigure}[b]{0.23\textwidth}
        \centering
        \includegraphics[width=\textwidth]{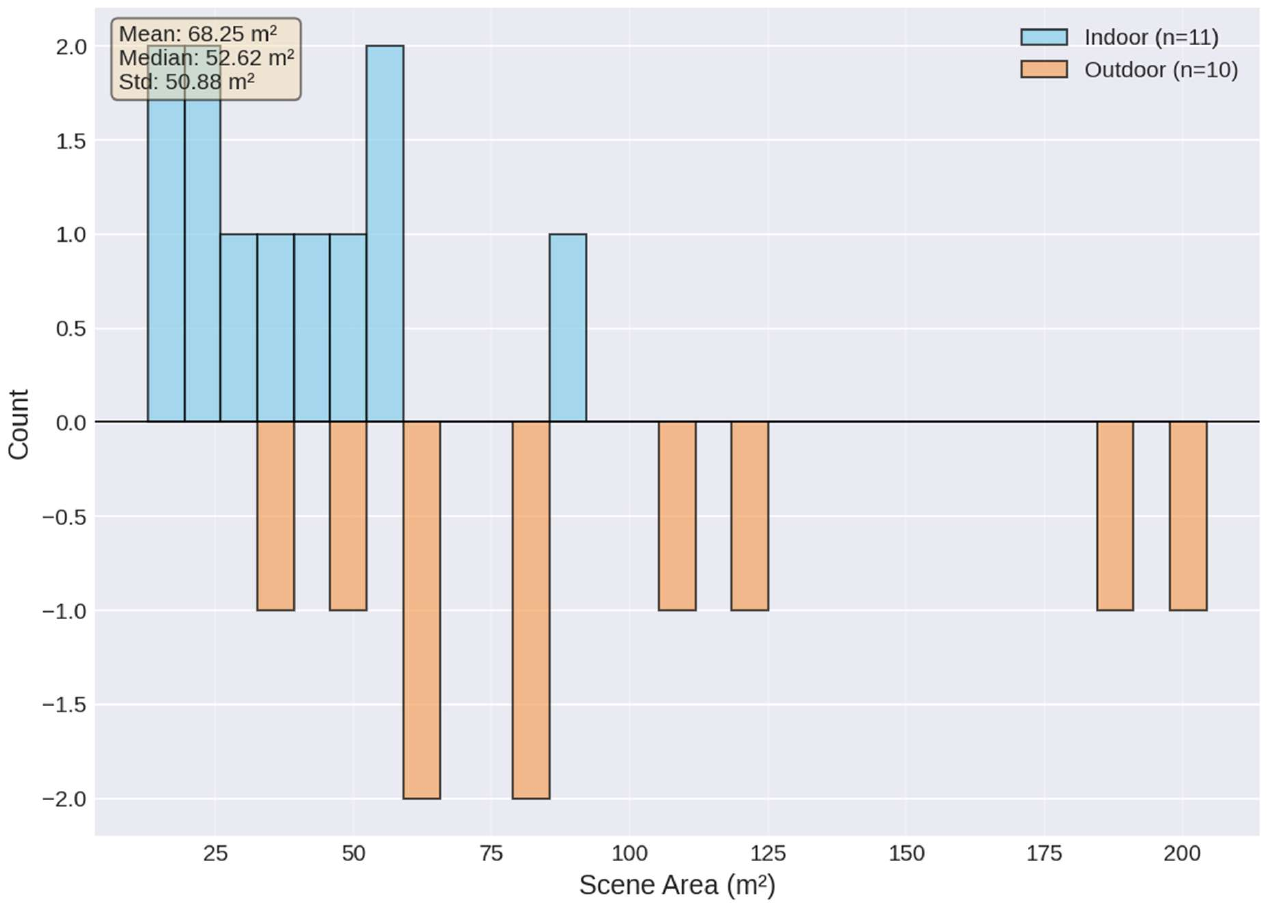}
        \caption{Scene Mesh Area Distribution.}
        \label{fig:stat3}
    \end{subfigure}
    \begin{subfigure}[b]{0.23\textwidth}
        \centering
        \includegraphics[width=\textwidth]{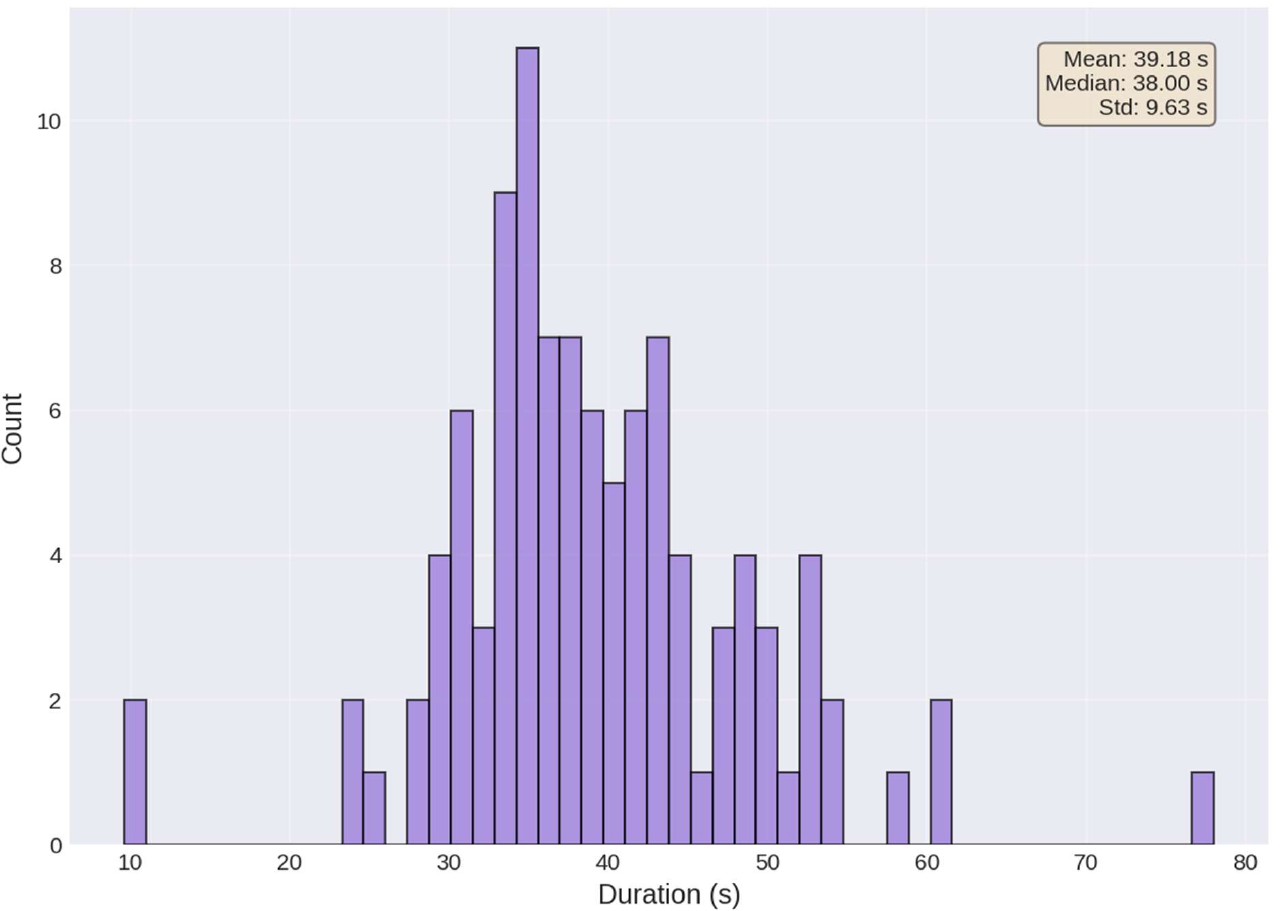} 
        \caption{Sequence Length Distribution.}
        \label{fig:stat4}
    \end{subfigure}
    \caption{Statistical information of collected dataset.}
    \label{fig:stats}
\end{figure}

\begin{figure*}[!ht]
    \centering
    % 设置占位框的宽度和高度
    \includegraphics[width=1\linewidth]{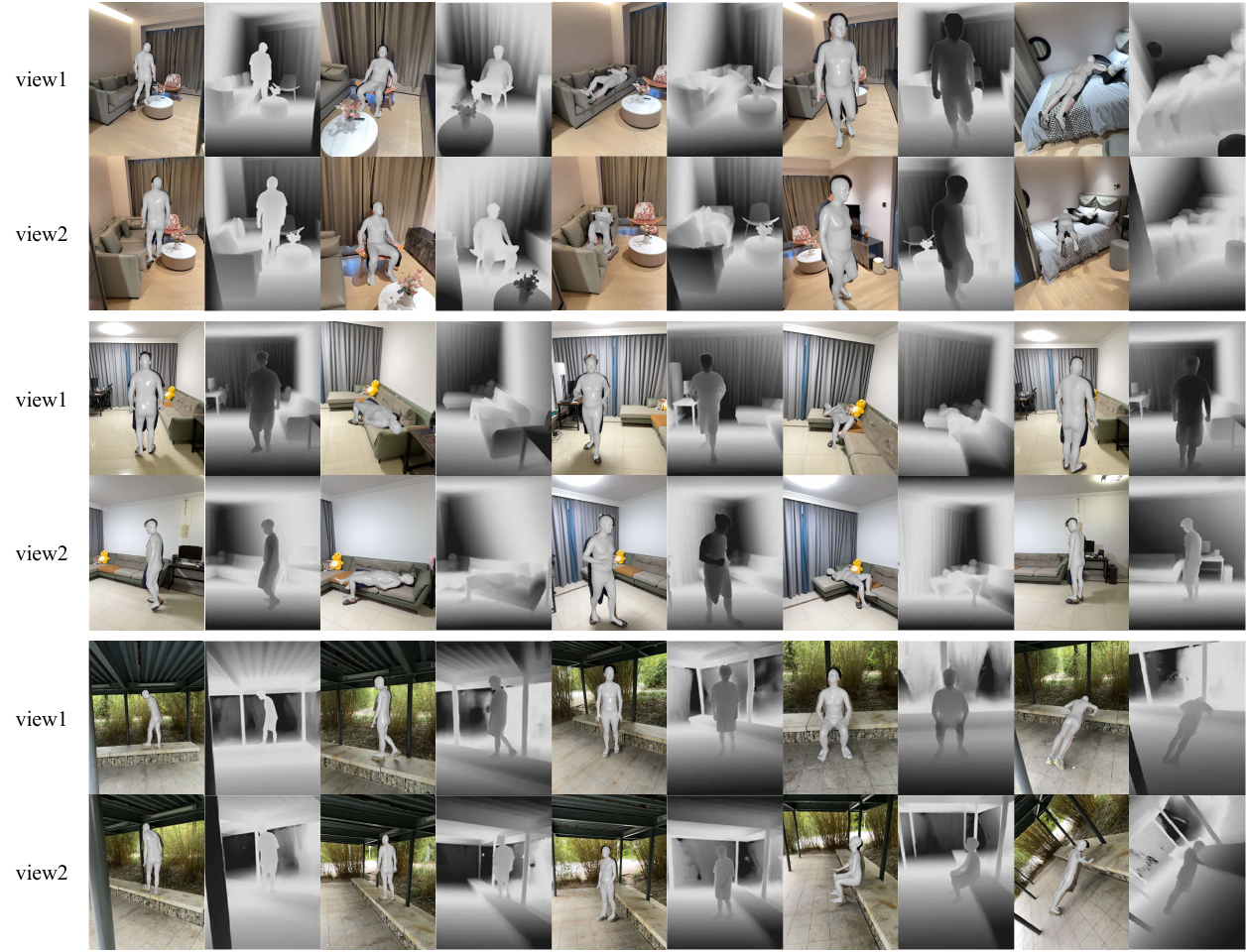}
    \vspace{-5pt}
    \label{fig:camera_space_demo}
    \caption{\textbf{Rendered SMPL and depth images of the captured dataset in camera space.} }
\end{figure*}

\mypara{Evaluation}
The evaluation of the Support task focuses on the agent’s ability to position its hands on the top surface of the target object and keep its feet close together. The key metric is the combined XY-plane distance and Z-axis deviation between the hands and the object’s top surface. The task is deemed successful if the hands are within predefined thresholds and the feet maintain adequate proximity for stability.

\section{More Details of Scene-Aware Imitation Policy}

\subsection{Representations}

\mypara{Character Proprioception.} The state $s$ describes the proprioception of the character's body, with features consisting of the relative positions of each link with respect to the root (designated to be the pelvis), their rotations expressed in quaternions, and their linear and angular velocities. All features are computed in the character's local coordinate frame, with the root at the origin and the x-axis along the root link's facing direction.

\mypara{Height Map.} To perceive the surrounding scene geometry, we utilize a local egocentric height map. This map is structured as an $11 \times 11$ grid spanning a $2\text{m} \times 2\text{m}$ area centered on the humanoid, resulting in a sampling interval of $0.2\text{m}$. The grid is defined within the character's local coordinate frame; consequently, the sampling points dynamically translate and rotate with the humanoid's movement and heading, consistently covering the immediate vicinity. The height values at these grid points are queried from a high-resolution underlying scene mesh (0.05m resolution) using nearest-neighbor interpolation. 

\mypara{Target States.} The target state $\hat{q}$ encodes the desired future motion of the character. It is constructed by sampling a short trajectory segment from the dataset spanning three consecutive future time steps: $T, T+1$, and $T+2$. For each time step, the state comprises the positions, rotations, linear velocities, and angular velocities of all body links. All features are transformed from the world frame into the simulated character's local coordinate frame. This local frame is defined with the character's root located at the origin and the x-axis aligned with the root link's facing direction.

\mypara{Action.} Our simulated humanoid is constructed based on the SMPL body model, comprising 23 controllable joints. Each joint possesses 3 degrees of freedom (DoF), and we employ a Proportional-Derivative (PD) controller for each DoF. Consequently, the action $a \in \mathbb{R}^{69}$ generated by the policy specifies the target orientations for these PD controllers.

\subsection{Reward}

To encourage the character to closely reproduce the reference motion while maintaining motion naturalness, our reward function $r_t$ is composed of two terms: a tracking reward $r^{\text{track}}_t$ and a jitter penalty $r^{\text{smooth}}_t$. The tracking reward incentivizes the policy to minimize the kinematic error between the simulated character and the reference motion. The jitter penalty is introduced to suppress abnormal shaking generated when the character interacts with objects, which may be induced by instabilities in the physics simulation. The total reward is defined as:
\begin{equation}
r_t = r^{\text{track}}_t - r^{\text{smooth}}_t.
\end{equation}
The tracking reward $r^{\text{track}}_t$ is computed as the weighted sum of exponential differences across all humanoid links:
\begin{equation}
    \begin{aligned}
        r^{\text{track}}_t = & \ w_{\text{jp}} \exp\left(-100 \| \hat{\boldsymbol{p}}_t - \boldsymbol{p}_t \|^2\right) \\
        & + w_{\text{jr}} \exp\left(-10 \| \hat{\boldsymbol{q}}_t \ominus \boldsymbol{q}_t \|^2\right) \\
        & + w_{\text{jv}} \exp\left(-0.1 \| \hat{\boldsymbol{v}}_t - \boldsymbol{v}_t \|^2\right) \\
        & + w_{\text{j}\omega} \exp\left(-0.1 \| \hat{\boldsymbol{\omega}}_t - \boldsymbol{\omega}_t \|^2\right),
    \end{aligned}
\end{equation}
where the equation penalizes the differences in translation $\boldsymbol{p}$, rotation $\boldsymbol{q}$, linear velocity $\boldsymbol{v}$, and angular velocity $\boldsymbol{\omega}$ for all rigid body links of the humanoid between the simulation and the reference. The jitter penalty penalizes the magnitude of the difference between consecutive actions, defined as:
\begin{equation}
    r^{\text{smooth}}_t = \| \boldsymbol{a}_t - \boldsymbol{a}_{t-1} \|^2,
\end{equation}
where $\boldsymbol{a}_t$ and $\boldsymbol{a}_{t-1}$ denote the action at the current and previous time steps, respectively. By minimizing the rate of change of the actions, the policy is incentivized to generate continuous and stable control trajectories, thereby reducing jittery behaviors.

\section{More Details of Captured Dataset Used in Main Paper}
We collected data from 23 scenes, each with a high-precision mesh, 104 sequences, and approximately 200,000 video frames. Each frame is accompanied by corresponding depth maps, segmentation masks, camera trajectories, and human parameters(bounding boxes, 2D keypoints, SMPL parameters).

In \cref{fig:stat1}, we present the distribution of camera trajectory lengths, which range from 4 meters to over 30 meters. In \cref{fig:stat2}, the human trajectory length distribution is shown, with performers moving between 5 meters and over 30 meters. \Cref{fig:stat3} illustrates the scene mesh area distribution. Indoor scenes are relatively smaller, ranging from 20 to 90 square meters, while outdoor scenes can be as large as 200 square meters. Finally, in \cref{fig:stat4}, we show the sequence length distribution, where most sequences have durations ranging from 30 to 60 seconds.

\subsection{Qualitative Demonstrations}
\begin{figure*}[!ht]
    \centering
    % 设置占位框的宽度和高度
    \includegraphics[width=1\linewidth]{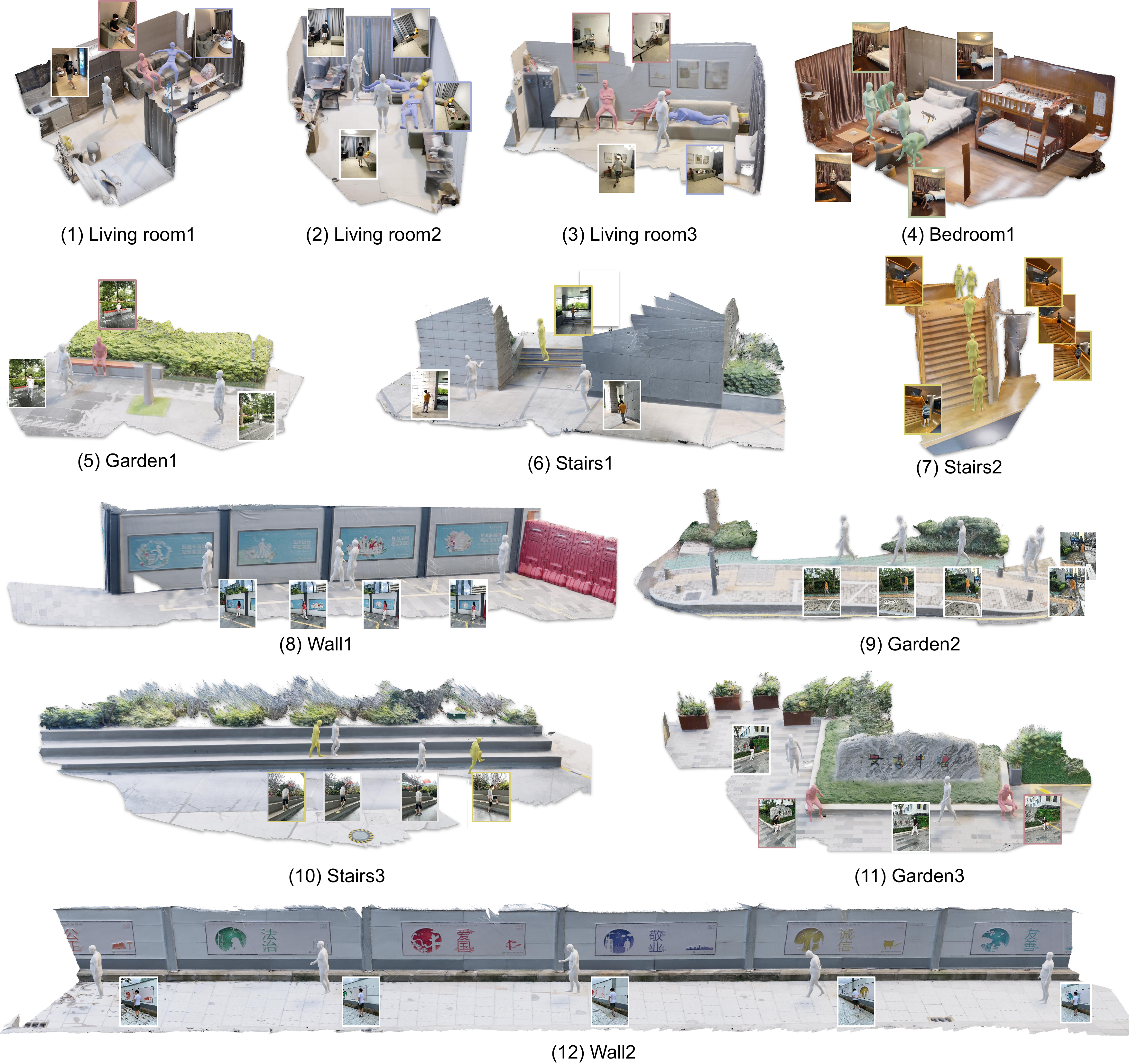}
    \label{fig:world_space_demo}
    \caption{\textbf{3D demo of the captured dataset.} }
    \vspace{-12pt}
\end{figure*}

We show camera space results in \cref{fig:camera_space_demo} and world space results in \cref{fig:world_space_demo}

